\documentclass{article}

\usepackage{arxiv}

\usepackage[utf8]{inputenc} 
\usepackage[T1]{fontenc}    
\usepackage{hyperref}       
\usepackage{url}            
\usepackage{booktabs}       
\usepackage{amsfonts}       
\usepackage{nicefrac}       
\usepackage{microtype}      

\usepackage{graphicx} 
\DeclareGraphicsExtensions{.pdf} 

\usepackage{amsmath} 

\usepackage[short]{optidef} 

\usepackage[ruled,linesnumbered]{algorithm2e} 

\usepackage{caption} 

\usepackage{bm} 

\usepackage{times} 
\usepackage{color,soul} 
\usepackage{color} 

\usepackage{amssymb} 
\newcommand{\quickwordcount}{%
	\immediate\write18{texcount -1 -sum -merge \jobname.tex > \jobname-words.sum }%
	\input{\jobname-words.sum} words%
}

\usepackage{amsthm} 

\newtheorem{assumption}{Assumption}
\newtheorem{remark}{Remark}
\theoremstyle{definition}
\newtheorem{theorem}{Theorem}

\newcommand{\mytilde}{\raise.17ex\hbox{$\scriptstyle\mathtt{‌​\sim}$}}
\newcommand{\myqed}{\tag*{$\square$}}

\renewcommand\footnotemark{}

\title{A Versatile Data-Driven Framework for Model-Independent Control of Continuum Manipulators Interacting with Obstructed Environments with Unknown Geometry and Stiffness}

\author{
  Farshid Alambeigi\textsuperscript{\textdagger} \\
  Department of Mechanical Engineering \\
  University of Texas at Austin \\
  Austin, TX 78712 \\
  \texttt{farshid.alambeigi@austin.utexas.edu} \\
   \And
  Zerui Wang\textsuperscript{\textdagger} \\
  Cornerstone Robotics Limited \\
  Shatin, Hong Kong SAR \\
  \texttt{zerui.j.wang@gmail.com} \\
   \And
  Yun-Hui Liu \\
  Department of Mechanical and Automation Engineering \\
  The Chinese University of Hong Kong \\
  Shatin, Hong Kong SAR \\
  \texttt{yhliu@mae.cuhk.edu.hk} \\
   \And
  Russell H. Taylor \\
  Laboratory for Computational Sensing and Robotics \\ 
  Johns Hopkins University \\
  Baltimore, MD 21218.\\
  \texttt{rht@jhu.edu} \\
   \And
  Mehran Armand \\
  Applied physics Laboratory \\
  Johns Hopkins university \\
  Laurel, MD 20723. \\
  \texttt{mehran.armand@jhuapl.edu} \\
  \thanks{\textsuperscript{\textdagger}Farshid Alambeigi and Zerui Wang are co-first authors in this work.}
}

\begin{document}
\maketitle

\begin{abstract}
	This paper addresses the problem of controlling a continuum manipulator (CM) in free or obstructed environments with no prior knowledge about the deformation behavior of the CM and the stiffness and geometry of the interacting obstructed environment.
	We propose a versatile data-driven priori-model-independent (PMI) control framework, in which various control paradigms (e.g. CM's position or shape control) can be defined based on the provided feedback.
	This optimal iterative algorithm learns the deformation behavior of the CM in interaction with an unknown environment, in real time, and then accomplishes the defined control objective.
	To evaluate the scalability of the proposed framework, we integrated two different CMs, designed for medical applications, with the da Vinci Research Kit (dVRK).
	The performance and learning capability of the framework was investigated in 11 sets of experiments including PMI position and shape control in free and unknown obstructed environments as  well as during manipulation of an unknown deformable object. We also evaluated the performance of our algorithm in an ex-vivo experiment with a lamb heart.
	The theoretical and experimental results demonstrate the adaptivity, versatility, and accuracy of the proposed framework and, therefore, its suitability for a variety of applications involving continuum manipulators.
\end{abstract}


\section{Introduction}
Continuum Manipulators (CM) can improve robot manipulation due to their added dexterity and safety as compared to conventional rigid link robots \cite{hirose2004biologically,trivedi2008soft}.
These features make CMs a viable candidate  for a variety of applications especially in medical interventions where a CM, due to its compliance, may interact safely with a deformable or rigid tissue with unknown deformation behavior \cite{camarillo2004robotic}.
The conformable structure of CMs, however, introduce additional challenges for accurate modeling and control of these robots.
These challenges will become more significant when the CM is working in a congested and obstructed environment \cite{yip2014model,yip2016model,george2017learning}.
In such environments, kinematics and dynamics of the CM differ from its free motion due to its undetermined interactions with obstacles.
Adaptive and intelligent control approaches are, therefore, necessary to maneuver the CM into a congested environment with unmodeled obstacles.

\subsection{Prior Work}
	To overcome the difficulties associated with the control of the  end-effector position and shape of the CM, researchers generally have implemented two types of approaches: priori-model-dependent (PMD) and priori-model-independent (PMI) control strategies.
	In the PMD approaches, a control strategy is implemented based on an obtained fixed and off-line kinematics- or dynamics-based model of the CM (e.g. \cite{camarillo2008mechanics,webster2006nonholonomic,rucker2011statics,rucker2010equilibrium,xu2010analytic}).
	However, robustness and performance of these analytical approaches are highly dependent on the accuracy of the developed model, which in turn, is a function of  simplifying assumptions made during the modeling step \cite{falkenhahn2017dynamic}.
	Literature documents various PMD modeling approaches and the parameters affecting the CM behavior including:
	deformation behavior (i.e. constant curvature \cite{webster2010design} or non-constant curvature behavior \cite{murphy2014design,mahl2014variable});
	mechanical structure (e.g. precurved concentric tubes \cite{mahvash2011stiffness,webster2009mechanics}, multi-backbone continuum robots \cite{simaan2005snake,simaan2009design,xu2010analytic}, and concentric tubes with lateral notches \cite{murphy2014design,gao2017mechanical});
	actuation mechanism (e.g. tendons \cite{murphy2014design,yip2014model}, pneumatic actuators \cite{mahl2014variable}, and shape memory alloys \cite{crews2012design});
	physical phenomenons such as hysteresis- caused by the elastic properties of the CM's constitutive material;
	friction (e.g. caused by actuation tendons); slack in the actuation mechanism \cite{camarillo2008mechanics,melingui2017adaptive,gao2017mechanical}, etc.
	Of note, neglecting any of these parameters may reduce the performance of a PMD control algorithm.
	
	Although most of the mentioned PMD control studies have been focused on free bending motion of the CM, some researchers have taken into account the effect of external disturbances in a constrained environment.
	For instance, \cite{bajo2012kinematics} studied a kinematics-based model identifying the contact point between a CM and a fixed obstacle in an unknown environment.  \cite{rucker2011statics} used a static CM deflection model and implemented an Extended Kalman filter to estimate the contact force. 
	
	PMI control approaches, however, do not rely on a fixed CM model and use adaptive methods to compensate undesired effects, caused by inaccurate modeling or environment disturbances, and estimate the CM behavior in real-time.
	\cite{yip2014model} documented one of the first works in the priori-model-independent control of continuum manipulators interacting with an obstructed environment.
	They used an optimal control strategy, called "model-less" control method, on a tendon-driven CM and guided their robot along a planar predefined trajectory in a constrained environment.
	\cite{yip2014model} have mentioned some limitations for their  approach.
	First, they used backward differencing of measurements to estimate Jacobian of their robot, resulting in a lag relative to the true Jacobian of the CM.
	This means that their algorithm uses the estimated Jacobian of the previous step for the current step, which may differ from the true Jacobian.
	Further, their method was sensitive to the measurement signal noise and using filters added to the lag of the estimation.
	Second, to estimate the Jacobian, their model-less method solves an under-constrained optimization problem minimizing the change in the Frobenius norm of the estimated Jacobian.
	However, there is more than one Jacobian solution that satisfies the defined constraints, which may not return the optimal solution and cause some misalignment in estimation of the Jacobian in each iteration.
	Third, using the proposed controller, the CM may get stuck in a singularity.
	This happens when the CM interacts with an obstructed environment via its end-effector (tip), where the estimation of the Jacobian with the proposed method becomes ill-conditioned.
	To solve the third issue, \cite{yip2016model} implemented a model-less hybrid position/force control and navigated their CM under body and tip constraints.
	However, the aforementioned first and second problems were not addressed in this work.
	They evaluated the performance of their approach in a constraint-free environment.
	
	Machine learning-based PMI controllers are another alternative solution for controlling the CMs.
	The use of neural networks for compensating unknown kinematics and dynamics of a CM, without presence of external disturbances, was studied by many researchers (e.g. \cite{melingui2015adaptive,giorelli2015neural,braganza2007neural}).
	As another  example, \cite{melingui2017adaptive} implemented an adaptive support vector regressor method to address position control of a CM in a free environment.
  	In a recent work, \cite{george2017learning} trained their neural network to move their continuum soft manipulator in an unstructured environment with external disturbances.
    A machine learning-based approach was also implemented to develop a dynamic model and a trajectory optimization method for predictive control of a particular soft robotic manipulator \cite{thuruthel2017learning}.
	In these studies, however, the networks were trained for a particular type of CM and environment and cannot be generalized for different types of CMs and environments.
	Also, similar to other machine learning approaches, the performance of this PMI off-line approach is highly dependent on the training data set obtained by cumbersome identification experiments, which may result in a poor local minimum solution.
	A reinforcement learning technique was implemented in \cite{engel2006learning} to address the problem of reaching a point by a simulated multi-segmented dynamical planar model of an octopus arm.
	In this study, a Hidden Markov Model was implemented  and solved using a nonparametric Gaussian temporal difference learning algorithm.
	\cite{silver2014deterministic} demonstrated that similar problem can be solved in the context of continuous action spaces using an actor-critic reinforcement learning approach. However, both of these studies performed in simulations and suffer from demanding real-time computational costs for generating solutions in practice \cite{george2018control}.
    
    Visual servoing is another approach, which can be utilized for PMD and PMI control of continuum manipulators.
    In a PMD approach (i.e. calibrated visual servoing), similar to other model-dependent approaches, a kinematics or dynamics model of the CM together with intrinsic and extrinsic camera parameters are required.
    Recently, \cite{kudryavtsev2018eye} utilized a PMD visual servoing approach to control a concentric tube robot in a free environment.
    Similar to other PMD approaches, those authors focused on control of the CM in a free environment and did not study the effect of unknown external disturbances on the performance of their algorithm.
    On the other hand, uncalibrated visual servoing  is a PMI control approach, which does not rely on robot kinematics and camera parameters as a priori information.
    Uncalibrated visual servoing was utilized for controlling conventional rigid-link robotic manipulators in a free environment without the presence of disturbances (e.g. \cite{piepmeier2004uncalibrated}). 

	\subsection{Our Contribution}
    We propose a versatile framework for Data-Driven PMI (DD-PMI) control of continuum manipulators.
    The algorithm extends the potential and application of  uncalibrated visual servoing approach \cite{piepmeier2004uncalibrated} to the control of CMs.
    The presented method not only can be used with various sensing modalities (e.g. camera and embedded sensors) but it also can be utilized to simultaneously learn the deformation behavior of the CM and the unknown  environment.
	The performance and scalability of the DD-PMI method are investigated using two different types of CMs with 3 degrees of freedom (DoF): a 3mm continuum manipulator especially designed and fabricated from nitinol tube for head and neck surgeries \cite{coemert2016development}, and a 5mm Endowrist Instrument (Intuitive surgical, Inc., California, USA).
	We integrated the two CMs with the da Vinci Research Kit (dVRK) \cite{kazanzides2014open} and performed several experiments in environments including rigid and deformable obstacles with different geometries.
	Specific contributions of this study considering the mentioned setup include:
	\begin{enumerate}
		\item Design of a versatile DD-PMI framework addressing the limitations of prior PMI control studies.
		\item Theoretical proof for the optimality of the estimated deformation behavior using the proposed learning method. 
		\item Implementing both DD-PMI CM shape and end-effector position control.
		\item Experimental evaluation of the DD-PMI framework in the case of both CM body and end-effector interactions with 6 different unknown obstructed environments without using a force sensor.
		\item Experimental demonstration of the DD-PMI algorithm robustness in the presence of the CM hysteresis, backlash, and unknown dynamic disturbances during manipulation of a deformable object.
	\end{enumerate}
	The remainder of this paper is organized as follows.
	In Section \ref{sec:modeling}, we present the mathematical models needed for model independent manipulation of CMs and discuss the deformation control framework.
	Experimental setup and evaluation procedures are presented in Section \ref{sec:experiment}.
	In Section \ref{sec:results}, results of three different sets of experiments are reported including DD-PMI control of both CMs in free and obstructed environments as well as the DD-PMI control during manipulation of an unknown deformable object.
	We also evaluated the DD-PMI algorithm in an ex-vivo experiment on a lamb heart and investigated the effect of parameters initialization on its convergence rate.
	In Section \ref{sec:discussion}, we discuss the results and Section \ref{sec:conclusion} concludes the paper.

	\section{Method}\label{sec:modeling}
	
    The goal of the DD-PMI algorithm is to autonomously manipulate an unmodeled continuum manipulator to satisfy a control objective in an unknown environment using an uncalibrated sensor.
	To achieve this objective, unknown deformation behavior of the CM interacting with an environment needs to be estimated and perceived in real time from the provided feedback by an uncalibrated sensor considering the following assumptions:
	
	\begin{assumption}
		\label{ass:visual-feedback}
		We use the visual feedbacks computed from the dVRK endoscope video to implement the DD-PMI method in the image plane.
		We assume that this visual feedback is always available from a camera, and we can measure the CM's deformation visually in real-time. The camera parameters (both extrinsic and intrinsic), however,  are not known and/or calibrated with respect to the robot(s). Also, this sensing can be noisy.
	\end{assumption}
	
	\begin{remark}
		\label{rem:feedback modalities}
		For the proposed approach, other imaging modalities such as ultrasound or optical tracker can substitute the endoscope video used for this study.
		Additionally, if line-of-sight is an issue, we can obtain the required feedback from embedded sensors such as Fiber Bragg Grating (FBG) optical \cite{liu2015shape} or magnetic sensors \cite{mahvash2011stiffness}.
	\end{remark}
	
	\begin{assumption}
		\label{ass:controllability and obsrvability}
		The proposed algorithm enables defining various control objectives (e.g. the control of the position of any point(s) on the body of the CM and its S-shape control) provided that 1) there exist(s) feedback(s) from all the desired point(s) of interest on the CM; 2) there is at least one possible physical solution for the desired control problem considering the CM actuation mechanism.
	\end{assumption}
	
	\begin{remark}
		\label{rem:identification of cm}
		There is no need to identify the CM's behavior prior to its manipulation by the proposed algorithm.
		Moreover, the CM's structure or actuation system can have unknown friction, hysteresis, slack, or backlash.
	\end{remark}
	
	\begin{remark}
		\label{rem:type of cm}
		We used two types of CM developed for medical applications. However, the proposed PMI control method can be implemented to various types of continuum manipulators and soft robots. 
	\end{remark}

	\begin{figure}[!t]
		\centering
		\includegraphics[width=.5\linewidth]{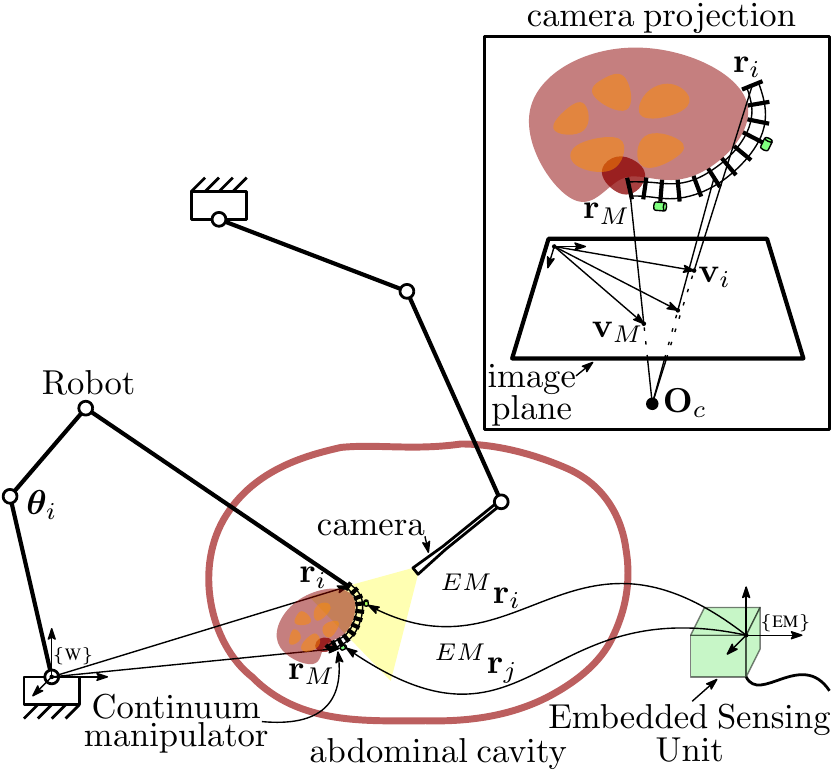}
		\caption{Conceptual illustration of the DD-PMI control of continuum manipulators in an obstructed environment using the visual feedback or an embedded sensing unit (e.g. FBG or magnetic sensors).
		In this method, $M$ sensing points on the surface of the CM are used to either control their positions or control the robot's shape, without having prior knowledge about the deformation behavior of the CM in interaction with the obstructed environment.
		In this figure, $\mathbf{r}_{i} \in \mathbb{R}^{3}$ represents the position of the $i$th feature point in the Cartesian frame $\{W\}$ or the embedded sensing unit frame $\{EM\}$, and $\mathbf{v}_{i} \in \mathbb{R}^{2}$ represents the position of the corresponding sensing point in the image plane of the visual sensing unit.
		Also, $\mathbf{O}_{c}$ denotes the camera optical center.}
		\label{fig:illustration}
	\end{figure}
	
	\subsection{Continuum Manipulator Jacobian}
	Consider $M$ points on the surface of the CM demonstrating locations we are either interested in controlling their position or using them to control the robot's shape.
	As shown in Fig. \ref{fig:illustration}, $\mathbf{r}_{i}(t) \in \mathbb{R}^{3}$ represents the position of the $i$th point in the Cartesian space:

	\begin{align}
	    \mathbf{r}_{i}(t) =
	    \begin{bmatrix}
	        x_i(t) & y_i(t) & z_i(t)
	    \end{bmatrix}^{\intercal}
	    \nonumber
	\end{align}
	where $i \in \{1, ..., M\}$. 
	
	Now, define vector of actuation inputs ${\bm{\theta}}(t) \in \mathbb{R}^{n}$ and for the ease of presentation stack the coordinates of all of the points in a long vector $\mathbf{r}(t) \in \mathbb{R}^{3M}$ as the following:
	\begin{equation} \label{eq:stackedvectorr}
	    \mathbf{r}(t)=
	    \begin{bmatrix}
	        \mathbf{r}^{\intercal}_1(t) & \mathbf{r}^{\intercal}_2(t) & ... & \mathbf{r}^{\intercal}_M(t)
	    \end{bmatrix}^{\intercal}
	    \nonumber
	\end{equation}
	
	Considering Assumption \ref{ass:controllability and obsrvability}, if the position of the $i$th point $\mathbf{r}_i(t)$ can be (locally) expressed as a smooth function of the vector of CM actuation inputs $\bm{\theta(t)}$, (i.e. $\mathbf{r}_i(t) = \mathbf{r}_i(\bm{\theta}(t)):\mathbb{R}^{n} \mapsto \mathbb{R}^{3}$), we can then locally model the interactive CM deformation behavior between displacements of the robot(s) actuation system and displacements of the points on the CM body as the following:

	\begin{align} \label{eq:interactive_deformation_jacobian_matrix}
	    \frac{d}{dt}\mathbf{r}(t) =
	    \underbrace{
		    \frac{\partial \mathbf{r}} {\partial \bm{\theta}}(\bm{\theta}(t))
	    }_{\mathbf{J}_{cm}(\bm{\theta}(t))}
	    \frac{d}{dt} \bm{\theta}(t)
	\end{align}
	where ${\mathbf{J}_{cm}(\mathbf{r}(t))}\in \mathbb{R}^{3M \times n}$ is the CM Jacobian matrix mapping variations of the actuation inputs to the related task space displacements of the points on the CM.
	
	Of note, considering (\ref{eq:interactive_deformation_jacobian_matrix}), for an infinitesimal time period $\Delta t$, the changes in the position(s) of the points $\Delta\mathbf{r}(t)$ can be related to the actuation input changes $\Delta\bm{\theta}(t)$ as:

	\begin{align} \label{eq:estimation Jcm}
	    \Delta\mathbf{r}(t) \approx {\mathbf{J}_{cm}}(\bm{\theta}(t)) \Delta\bm{\theta}(t)
	\end{align}
	
	\subsection{Feedback Feature Vector Function}
	
	In this study, we use the visual feedback provided by an uncalibrated camera (as our sensor) to implement the DD-PMI control algorithm in the image plane.
	However, as mentioned in Assumption \ref{ass:visual-feedback}, other sensors can also be used to provide the required feedback in their local coordinate frames.
	Considering the provided feedback, we can mathematically define $N$ desired control objectives (e.g. overlaying CM's end-effector position to a predefined position or giving a particular curvature to the CM) using the \textit{Feedback Feature Vector} (FFV) function $\bm{\Gamma}(t) \in \mathbb{R}^{N}$.
	
	To this end, considering the visual feedback, we first need to relate the coordinates of the feature points on the CM in the image plane $\mathbf{v}_{i}(t) \in \mathbb{R}^{2}$ to their corresponding position in the Cartesian space $\mathbf{r}_{i}(t) \in \mathbb{R}^{3}$ using the pin-hole perspective projection model without distortion \cite{forsyth2002computer}:
	
	\begin{align}
	    \mathbf{v}_{i}(t)
	    & =
	    \begin{bmatrix}
	        p_i(t) & q_i(t)
	    \end{bmatrix}^{\intercal}
	\end{align}
	\begin{align} \label{eq:pin-hole perspective projection}
	    \begin{bmatrix}
	        \mathbf{v}_{i}(t)\\
    	    1
    	\end{bmatrix}
    	=
    	\underbrace{\frac{1}{^{cam}z_i(t)} \mathbf{P}}_{\mathbf{J}_{p_i}(t)}
    	\begin{bmatrix}
    	    \mathbf{r}_{i}(t)\\
    	    1
	    \end{bmatrix}
	\end{align}
	where $p_i(t)$ and $q_i(t)$ are the coordinates of the feature points in the image frame, ${^{cam}z_i(t)}$ is the third element of the feature point position $\mathbf{r}_{i}(t)$ expressed in the camera frame and $\mathbf{P}$ is the perspective projection matrix, which contains the information of intrinsic and extrinsic camera parameters.
	Of note, in this paper, besides unknown kinematics behavior of the CM, we assume the projection Jacobian matrix $\mathbf{J}_{p_i}(t) \in \mathbb{R}^{3 \times 4}$ is unknown.
	Using (\ref{eq:pin-hole perspective projection}), we can write the FFV as a smooth function of the feature points coordinates in the image plane $\bm{\Gamma}(\mathbf{v}(t) $, where $\mathbf{v}(t) \in \mathbb{R}^{2M}$ is the stack of all the feature points positions in the image plane:
	
	\begin{align} \label{eq:v_stacked_vector}
	    \mathbf{v}(t)
	    & =
	    \begin{bmatrix}
	        \mathbf{v}^{\intercal}_1(t) & \mathbf{v}^{\intercal}_2(t) & ... & \mathbf{v}^{\intercal}_M(t)
	    \end{bmatrix}^{\intercal}
	\end{align}
	
	As two examples of the FFV, in this paper, we consider the following functions for evaluation of the DD-PMI method.
	
	\subsubsection{Overlaying FFV Function:}
	Considering Assumption (\ref{ass:controllability and obsrvability}) and stacked feature vector (\ref{eq:v_stacked_vector}), we use the following FFV to overlay position of one or multiple feature points, on the CM, on to the desired predefined location(s) in the image plane $\mathbf{v}_d(t) \in \mathbb{R}^{2M}$:

	\begin{align} \label{overlaying ifv}
	    \bm{\Gamma}(t)=\mathbf{v}(t)
	\end{align}
	
	\subsubsection{Curvature Control FFV Function:}
	We use this FFV to control the shape (i.e. curvature) of the CM in a free environment without the presence of external disturbances.
	For this function, we need three feature points ($\mathbf{v}_1,\mathbf{v}_2$, and $\mathbf{v}_3 $) on the CM to define a curve that passes through these points with the curvature $\kappa \in \mathbb{R}$ obtained by the following relation \cite{kimberling1998triangle}:

	\begin{align} \label{eq:curvature k}
	    \bm{\Gamma}(\mathbf{v}(t))
    	=
    	\kappa(t)
    	= 
    	\frac{2{\left\|(\mathbf{v}_1-\mathbf{v}_2)\times(\mathbf{v}_2-\mathbf{v}_3)\right\|}}
    	{{\left\| (\mathbf{v}_1-\mathbf{v}_2) \right\|} {\left\| (\mathbf{v}_2-\mathbf{v}_3) \right\|} {\left\| (\mathbf{v}_3-\mathbf{v}_1) \right\|}}
    \end{align}
	
	\subsection{The Feedback Feature Jacobian Matrix}
	
	Considering the definition of the FFV, the CM Jacobian, and the perspective projection model, we can relate the time derivative of the FFV to the CM actuation inputs as the following:

	\begin{align} \label{eq:combined_image_jacobian}
    	\frac{d}{dt}\bm{\Gamma}(\mathbf{v}(t))
    	& =
    	\frac{\partial \bm{\Gamma}}{\partial \mathbf{v}} (\mathbf{v}(t)) \frac{d}{dt}\mathbf{v}(t) \nonumber \\
    	& =
    	\underbrace{\frac{\partial \bm{\Gamma}}{\partial \mathbf{v}} (\mathbf{v}(t)) \frac{\partial \mathbf{v}}{\partial \mathbf{r}} (\mathbf{r}(t))}_{\mathbf{J}_{if}(\mathbf{r}(t))} \frac{d}{dt} \mathbf{r}(t) \nonumber \\
    	& = {\mathbf{J}_{if}(\mathbf{r}(t))}\underbrace{\frac{\partial \mathbf{r}} {\partial \mathbf{\theta}}(\mathbf{\theta}(t))}_{\mathbf{J}_{cm}(\mathbf{\theta}(t))} \frac{d}{dt} \mathbf{\theta}(t) \nonumber \\
    	& = \mathbf{J}_{if}\mathbf{J}_{cm}\frac{d}{dt} \mathbf{\theta}(t)
	\end{align}
	where we call ${\mathbf{J}_{if}(\mathbf{r}(t))}\in\mathbb{R}^{N \times 3M}$ ($n \ge N$) as the \textit{feedback feature Jacobian matrix}.
	This matrix relates the change of the feature point(s) positions to the change of the FFV.
	With a similar analogy used in (\ref{eq:estimation Jcm}), we can approximate (\ref{eq:combined_image_jacobian}) as the following:

	\begin{align} \label{eq:estimation comibined jacobian}
    	\Delta\mathbf{\Gamma}(t) \approx \underbrace{\mathbf{J}_{if}\mathbf{J}_{cm}}_{\mathbf{J}_{comb}} \Delta\bm{\theta}(t)
	\end{align}
	where $\mathbf{J}_{comb}\in \mathbb{R}^{N \times n}$, called \textit{combined Jacobian matrix}, is the multiplication of the CM and feedback feature Jacobian matrices.
	\begin{remark} \label{rem:unknown jacobians}
		In this paper, both $\mathbf{J}_{if}$ and $\mathbf{J}_{cm}$, and subsequently $\mathbf{J}_{comb}$, are considered as unknown Jacobian matrices, which need to be estimated in real-time.
		However, as mentioned, depending on the sensor used to provide feedback during DD-PMI control, we may need to only estimate $\mathbf{J}_{cm}$.
	\end{remark}
	
	\subsection{Data-Driven PMI Control Approach}
	
	As mentioned, the control objectives can be mathematically defined using the feedback feature vector function (e.g. (\ref{overlaying ifv}) and (\ref{eq:curvature k})).
	Hence, if the FFV converges to a predefined desired feature vector $\bm{\Gamma}_d$ then these objectives have been satisfied:
	
	\begin{align}
    	\label{eq:error of ifv}
    	\widetilde{\mathbf{\Gamma}}(t) = \bm{\Gamma}(t) - \bm{\Gamma}_d
	\end{align}
	where $\widetilde{\mathbf{\Gamma}}(t)\in \mathbb{R}^{N}$ defines the error vector of control objectives.
	
	So, considering (\ref{eq:estimation comibined jacobian}) and (\ref{eq:error of ifv}), we define the following optimization problem- subject to various constraints- to successfully achieve the control objectives:
	\begin{argmini}
		{\Delta\bm{\theta}(t)}{\left\| \mathbf{J}_{comb} \Delta\bm{\theta}(t) - \widetilde{\mathbf{\Gamma}}(t)\right\|^{2}_{2}}
		{\label{eq:optimmain}}{}
		\addConstraint{\mathbf{A}(t)\Delta\bm{\theta}(t)}{\le \mathbf{b}(t)}
		\addConstraint{\Delta\bm{\theta}(t)}{\le \Delta\bm{\theta}_{\mathrm{max}}}
		\addConstraint{\Delta\bm{\theta}(t)}{\ge \Delta\bm{\theta}_{\mathrm{min}}}
	\end{argmini}
	where matrix $\mathbf{A}(t) \in \mathbb{R}^{h \times n}$ and vector $\mathbf{b}(t)\in \mathbb{R}^{h}$ define $h$ linear inequality constraints.
	Further, the last two constraints impose bounds on the CM actuation inputs.
	Of note, these constraints can easily be defined in the task space and (or) the sensor space (e.g. image plane) using the appropriate Jacobian matrix.
	
	\subsubsection{Real-Time Estimation of the Deformation Jacobian:}
	
	As mentioned in Remark (\ref{rem:unknown jacobians}), in each time instant using the provided feedback, we need to estimate $\mathbf{J}_{comb}$ before solving the defined optimization problem in (\ref{eq:optimmain}).
	In fact, equation (\ref{eq:estimation comibined jacobian}) is a system of $N$ linear equations with $ N \times n$ unknowns, where the unknowns are the elements of $\mathbf{J}_{comb}$.
	Therefore, this equation is an under-determined equation, which does not uniquely determine all the components of $\mathbf{J}_{comb}$.
	To solve this problem, \cite{yip2014model} used a constrained optimization method, which tries to minimize the change in the Frobenius norm of the estimated Jacobian.
	However, there is more than one Jacobian solution that satisfies the defined constraints, which may not return the optimal solution and cause some misalignment in estimation of Jacobian in each iteration.
	To address this issue, we instead implement the \textit{Broyden's method} \cite{broyden1965class} to iteratively approximate $\mathbf{J}_{comb}$.
	This method is a recursive first-rank update rule used for solving under-determined nonlinear system of equations \cite{wright1999numerical}.
	For instance, this update rule is used for visual servo control of uncalibrated robotic systems \cite{piepmeier2004uncalibrated}.
	
	If we assume $\widehat{\mathbf{J}}_{comb}$ is a matrix approximating $\mathbf{J}_{comb}$,
	then, it should satisfy the following equation- called "\textit{Secant condition}"- to ensure both $\widehat{\mathbf{J}}_{comb}(t + \Delta t)$ and ${\mathbf{J}}_{comb}(t + \Delta t)$ demonstrate similar behavior along the direction $\bm{s_k}$ \cite{wright1999numerical}. In other words, given the same input $\bm{s_k}$, both estimated and real Jacobians return identical output $\bm{y_k}$.
	\begin{equation} \label{eq:secant condition}
    	\bm{y_k}(t)={\widehat{\mathbf{J}}_{comb}(t + \Delta t)} \bm{s_k}(t)
	\end{equation}
	where $\bm{s_k}(t)=\bm{\theta}(t+\Delta t)-\bm{\theta} (t)$ and $\bm{y_k}(t)=\bm{\Gamma}(t+\Delta t)- \Gamma(t)$.
	
	\begin{theorem} \label{TH:theorem1}
		Among all matrices $\widehat{\mathbf{J}}_{comb}$ satisfying the \textit{Secant} condition (\ref{eq:secant condition}), the matrix ${\widehat{\mathbf{J}}_{comb}(t + \Delta t)}$ defined by the following equation results in the smallest possible change in the Frobenius matrix norm ${\left\|\widehat{\mathbf{J}}_{comb}(t + \Delta t)-{\widehat{\mathbf{J}}_{comb}(t)}\right\|}$:
		
		\begin{align} \label{eq:broyden}
		& {\widehat{\mathbf{J}}_{comb}(t + \Delta t)} \nonumber \\
		& = {\widehat{\mathbf{J}}_{comb}(t)} +\beta \frac{\Delta \bm{\Gamma}(t)-{\widehat{\mathbf{J}}_{comb}(t)} \Delta\bm{\theta}(t)}{ \Delta\bm{\theta}(t))^{\intercal} \Delta\bm{\theta}(t)} \left(\Delta\bm{\theta}(t)\right)^{\intercal}
		\end{align}
		
		where $0 \leq \beta \leq 1$ is a constant parameter, controlling the rate of change of $\widehat{\mathbf{J}}_{comb}(t)$.
		In other words (\ref{eq:broyden}) is the solution of the following optimization problem:
		\begin{argmini}
			{{\widehat{\mathbf{J}}_{comb}(t + \Delta t)}}{\left\| \widehat{\mathbf{J}}_{comb}(t + \Delta t)-{\widehat{\mathbf{J}}_{comb}(t)} \right\|^{2}_{2}}
			{\label{eq:optim}}{}
			\addConstraint{\bm{y_k}(t)}{={\widehat{\mathbf{J}}_{comb}(t + \Delta t)} \bm{s_k}(t)}
			\nonumber
		\end{argmini}
	\end{theorem}
	
	\begin{proof}
		Let $\widehat{\mathbf{J}}_{comb}$ be any estimated matrix satisfying (\ref{eq:secant condition}), using the properties of Euclidean norm and (\ref{eq:broyden}) for any vector $\Delta\bm{\theta}$, we have:

		\begin{align}
    		&
    		{\left\| \widehat{\mathbf{J}}_{comb}(t + \Delta t) - \widehat{\mathbf{J}}_{comb}(t) \right\|}
    		\nonumber \\
    		= &
    		{\left\| \frac{\Delta \bm{\Gamma}(t) - {\widehat{\mathbf{J}}_{comb}(t) } \Delta\bm{\theta}(t)}{ (\Delta\bm{\theta}(t))^{\intercal} \Delta\bm{\theta}(t)} \left( \Delta\bm{\theta}(t)\right)^{\intercal}\right\| }
    		\nonumber \\
    		= &
    		{\left\| \frac{\widehat{\mathbf{J}}_{comb} - \widehat{\mathbf{J}}_{comb}(t)}{ (\Delta\bm{\theta}(t))^{\intercal} \Delta\bm{\theta}(t)} \Delta\bm{\theta}(t)( \Delta\bm{\theta}(t))^{\intercal}\right\|}
    		\nonumber \\
    		\le &
    		{\left\| \widehat{\mathbf{J}}_{comb} - {\widehat{\mathbf{J}} _{comb}(t)} \right\|}
    		{\left\| \frac{\Delta\bm{\theta}(t) (\Delta\bm{\theta}(t))^{\intercal}} {(\Delta\bm{\theta}(t))^{\intercal} \Delta\bm{\theta}(t)}\right\|}
    		\nonumber \\
    		= &
    		{\left\| \widehat{\mathbf{J}}_{comb} - {\widehat{\mathbf{J}}_{comb}(t)} \right\|}
		    \nonumber
		\end{align}
		
		In this proof, we used the fact that

		\begin{align}
    		\left\|
    		\frac{ \Delta\bm{\theta}(t) (\Delta\bm{\theta}(t))^{\intercal} }{(\Delta\bm{\theta}(t))^{\intercal} \Delta\bm{\theta}(t)}
    		\right\|
    		=
    		1
    		\nonumber
    	\end{align}
    		
    	since ${\Delta\bm{\theta}(t) \left (\Delta\bm{\theta}(t)\right)^{\intercal} }$ is symmetric: $\forall \bm{\nu} \in \mathbb{R}^{n}$
    
    	\begin{align}
    		\left\| \frac{ \Delta\bm{\theta}(t) (\Delta\bm{\theta}(t))^{\intercal} }{ (\Delta\bm{\theta}(t))^{\intercal} \Delta\bm{\theta}(t)} \right\|
    		& =
    		\underset{\| \nu \|=1}{\text{max}}
    		\frac{{\bm{\nu}^{\intercal} \Delta\bm{\theta}(t) (\Delta\bm{\theta}(t))^{\intercal} \bm{\nu}} }{ (\Delta\bm{\theta}(t))^{\intercal}\Delta\bm{\theta}(t)}
    		\nonumber \\
    		& =
    		\frac{ {\|\Delta\bm{\theta}(t)\|}^2}{\|\Delta\bm{\theta}(t)\|^2}
    		\nonumber 
    		= 1 \myqed
    		\nonumber
		\end{align}
	\end{proof}
	
	Therefore, considering Theorem \ref{TH:theorem1}, using (\ref{eq:broyden}) guarantees that the estimated Jacobian in each step approximately has identical behavior with the true Jacobian and does not show lag relative to its true value due to imposing the Secant condition.
	Further, it has the minimum possible change relative to the last estimated matrix, which results in a uniform behavior for the continuum manipulator and non-jerky motions.
	
	\subsubsection{DD-PMI Control Algorithm:}
	We summarize the DD-PMI control method as the following: first, we define the control objective(s) of the problem using appropriate FFV functions (e.g. (\ref{overlaying ifv}) and (\ref{eq:curvature k})) as well as their desired values $\bm{\Gamma}_d$.
	We also need to define the constraints of the optimization problem in (\ref{eq:optimmain}) and the error threshold for the algorithm $\epsilon$.
	Further, the estimated combined Jacobian should be initialized with a non-singular matrix $\widehat{\mathbf{J}}_{comb}(0)$ (e.g. Identity matrix) in (\ref{eq:broyden}).
	Of note, better initialization of this matrix makes its estimation faster. 
	
	After defining the problem and initializing the parameters, in each time instant $(t + \Delta t)$, given the estimated deformation Jacobian matrix $\widehat{\mathbf{J}}_{comb}(t)$ determined by (\ref{eq:broyden}) and the CM actuation control input $\Delta\bm{\theta}(t)$ calculated from (\ref{eq:optimmain}), the CM is moved.
	Then, the modified CM Jacobian $\widehat{\mathbf{J}}_{comb}(t + \Delta t)$ is calculated using the actual measured displacement of the feature point(s) on the CM (i.e. $\mathbf{v}_{i}(t)$) in the image plane, which considering Theorem (\ref{TH:theorem1}) will result in a reduced error compared to the time instant $t$.
	This algorithm iterates while a predefined error threshold $\epsilon$ is satisfied.
	Algorithm \ref{alg:pmi} summarizes these steps.
	
	\let\oldnl\nl
	\newcommand{\nonl}{\renewcommand{\nl}{\let\nl\oldnl}}
	\begin{algorithm}[!h]
		\DontPrintSemicolon
		\nonl\textbf{Initialization}:\;
		$\bm{\Gamma},\bm{\Gamma}_d \longleftarrow$ choose the Feedback Feature Vector function from (\ref{overlaying ifv}) or (\ref{eq:curvature k}) as well as their desired values\;
		$\mathbf{A}, \mathbf{b}, \Delta\bm{\theta}_{\mathrm{min}}, \Delta\bm{\theta}_{\mathrm{max}} \longleftarrow$ define constraints\;
		$\widehat{\mathbf{J}}_{comb}(0) \longleftarrow$ initialize the combined Jacobian matrix\; 
		$\epsilon \longleftarrow$ define the error threshold \;
		$\beta \longleftarrow$ define the rate of change of $\widehat{\mathbf{J}}_{comb}(0)$ in (\ref{eq:broyden})\;
		\nonl\textbf{Control loop}:\;
		\While{$\|\widetilde{\mathbf{\Gamma}}\| > \epsilon$}{
			$\Delta{\bm{\Gamma}} \longleftarrow$ update the error vector using (\ref{eq:error of ifv}) \;
			$\Delta\bm{\theta} \longleftarrow$ (\ref{eq:optimmain}) calculate the actuation control inputs \;
			Command the CM with the calculated $\Delta\bm{\theta}$\;
			$\Delta\mathbf{q} \longleftarrow$ update the vector of the feature points (\ref{eq:v_stacked_vector})\;
			$\widehat{\mathbf{J}}_{comb} \longleftarrow$ (\ref{eq:broyden}) update the combined Jacobian matrix\;
		}
		\caption{DD-PMI Control of CMs }
		\label{alg:pmi}
	\end{algorithm}

	\section{Experimental Setup} \label{sec:experiment}
	
\begin{figure}[!t]
		\centering
		\includegraphics[width=.5\linewidth]{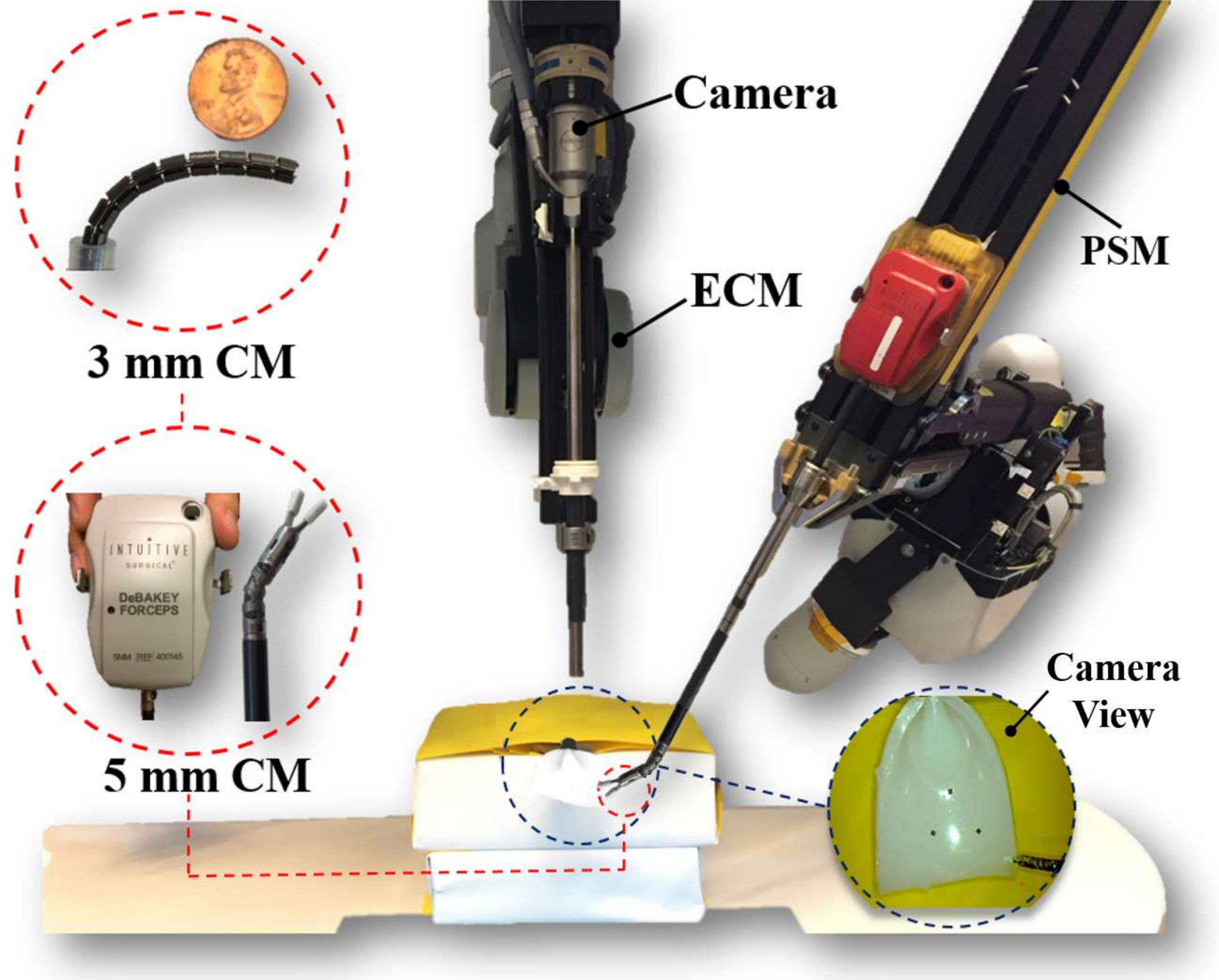}
		\caption{The setup used for the DD-PMI control of continuum manipulators using the dVRK system. We actuated a 5mm EndoWrist DeBakey Forcep and a custom-designed 3mm continuum manipulator with the da Vinci Instrument's actuation box to perform our experiments.}
		\label{fig:setup}
	\end{figure}

	To evaluate the DD-PMI method, we integrated two different types of continuum manipulators with the da Vinci Research Kit (Fig. \ref{fig:setup}).
	The following sections briefly describe our experimental setup and procedure as well as the considered evaluation measures.

	\subsection{The da Vinci Research Kit (dVRK)}
	
	The dVRK consists of mechanical parts from the da Vinci Classic Surgical System donated by Intuitive Surgical Inc., and open source electronics and software (cisst/SAW libraries) developed by researchers at Johns Hopkins University \cite{kazanzides2014open}.
	This surgical robot has three Patient Side Manipulators (PSMs), one Endoscopic Camera Manipulator (ECM), and two Master Tool Manipulators.
	For our evaluation experiments, we integrated the following CMs with one of the PSMs, and used an endoscope mounted on the ECM for obtaining the visual feedback (Fig. \ref{fig:setup}).
	
	\subsection{5 mm EndoWrist DeBakey Forceps}
	
	Integration of this da Vinci EndoWrist Instrument (Intuitive surgical, Inc., California, USA) with the dVRK, similar to human wrist articulation, provides 7 DoF at the instrument's distal end.
	Of note, the instrument solely has 4 DoF, which is controlled by its actuation box.
	Despite other multiport EndoWrist instruments, this type of forceps incorporate a ‘‘snake style’’ wrist to provide more dexterity to the surgeon (Fig. \ref{fig:setup}).
	The flexible part of the instrument is 14 mm long and can bend in two orthogonal planes.
	It also can have a rotational motion about the central axis of the instrument's rigid shaft.
	The fourth degree of freedom of the instrument is used for grasping objects.
	In this paper, we only use the 4 DoFs provided by the instrument to control the flexible part of this continuum instrument in free and obstructed environments.
	
	\subsection{3 mm Continuum Manipulator}
	
	This continuum manipulator with 3.3 mm outer diameter and 1.8 mm tool channel diameter has been designed to assist surgeons in treatment of petrous apex lesions \cite{coemert2016development}.
	It is made of nitinol with 40 mm working length and has an endoscope channel with a diameter of 0.7 mm (Fig. \ref{fig:setup}).
	To actuate this tendon-driven robot using the dVRK and obtain C- and S-shapes, we modified the da Vinci Instrument's actuation box \cite{coemert2016integration}.
	Utilizing this actuation box, the robot can be actuated in one plane and two directions and rotated around its rigid shaft axis.
	In contrast to the 5 mm instrument, the 3 mm manipulator shows hysteresis in its structure, due to the fabrication process and its delicate structure.
	It also has slack in its actuation mechanism.
	\begin{figure*}[!t]
		\centering
		\includegraphics[width=\linewidth]{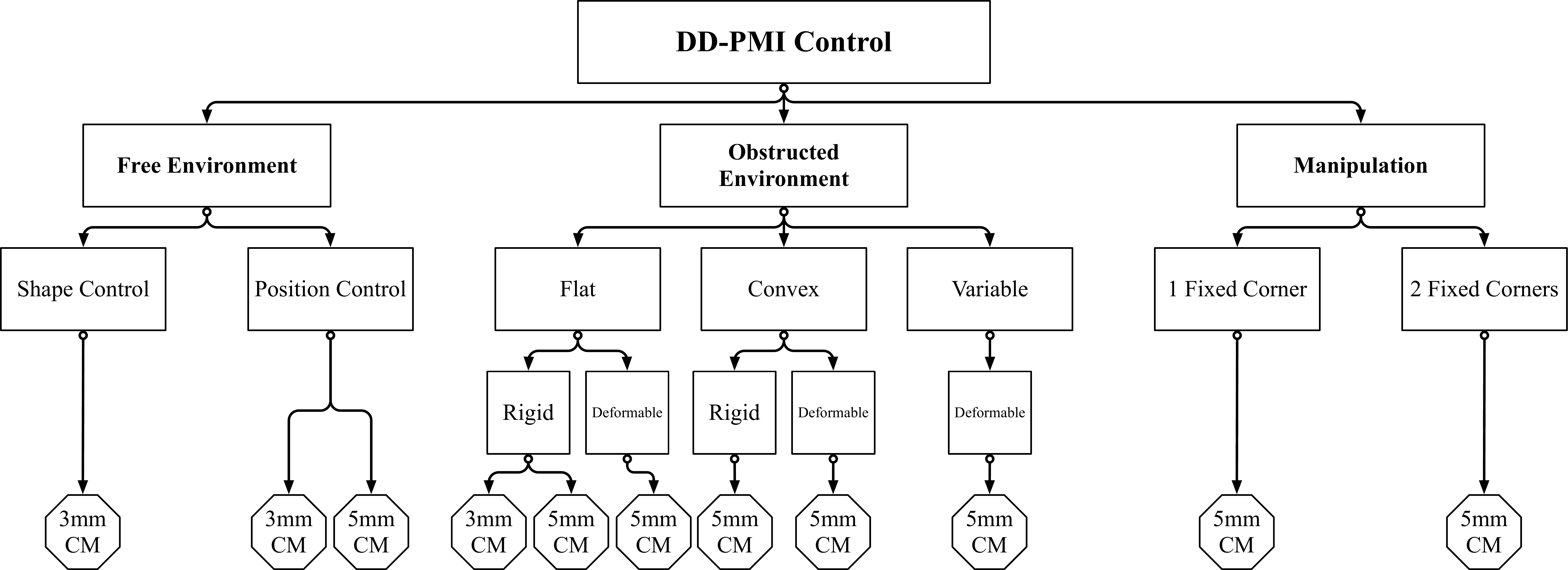}
		\caption{A summary of 11 experiments performed in this study to evaluate the proposed versatile DD-PMI control framework.
		We categorized the position and shape control experiments based on the geometry and stiffness of the environments as well as the type of the CM used to peform these experiments.}
		\label{fig:summary}
	\end{figure*}

	\subsection{Experimental Procedure and Evaluation Measures}
	
	Before performing each experiment, we first integrated the CM with the dVRK and assigned feature point(s) on the body of the CM.
	The endoscope position and orientation was arbitrarily chosen such that it provided enough field of view. Hence,   we are not calibrating the robots and camera before performing experiments.
	To track displacement of the feature point(s) using endoscope, we implemented a visual tracking algorithm with the Optical Flow package of the OpenCV library \cite{bradski2000opencv}, which implements the Lucas-Kanade algorithm \cite{baker2004lucas}.
	To control the movement of the CM using dVRK, we used cisst/SAW libraries in C++ and MATLAB (MathWorks, Inc.).
	We also used a MATLAB-ROS bridge \cite{quigley2009ros} to communicate between these environments.
	To solve (\ref{eq:optimmain}), we used an activeset algorithm \cite{gill1981practical}- i.e. the \textit{lsqlin} function in MATLAB- which is used for solving constrained linear least-squares problems. The rate of camera feedback was 25 Hz and the update rate of the total control loop (including the feedback and other calculations using MATLAB) was 15 Hz for all experiments.
	
	As mentioned in Assumption \ref{ass:visual-feedback}, the visual feedback provided by the vision system can be impeded by measurement noise.
	To mitigate this issue, we first filtered the measurements using the following first-order low-pass filter and then used them in (\ref{eq:broyden}) and (\ref{eq:optimmain}):

	\begin{align}
	    \dot{\mathbf{x}}_{f} = -\bm{\Lambda} (\mathbf{x}_{f} - \mathbf{x})
	\end{align}
	where $\mathbf{x}$ and $\mathbf{x}_{f}$ represent the original and filtered signals, respectively.
	The diagonal matrix $\bm{\Lambda} = \mathrm{diag}(\lambda_1, \lambda_2, \cdots, \lambda_k)$ allows us to adjust the dissipative properties of the filtered signal.
	
	To analyze the results of each experiment and the performance of the DD-PMI method, we used two measures as follows.
	To study the convergence for each experiment, we plotted the trajectory of the CM's end effector in the image plane, and calcualted the Euclidean distance between the desired target point and the end-effector position at each time step.
	We call this as the Euclidean distance error (EDE).
	We also applied the Yoshikawa's manipulability measure (YMM) $\Phi$ \cite{yoshikawa1985manipulability} to study the performance of the  \textit{Broyden} update rule in real-time estimation of the combined deformation Jacobian $\widehat{\mathbf{J}}_{comb}$:

	\begin{align} \label{eq:yoshikawa}
	    {\Phi} = \sqrt{det(\widehat{\mathbf{J}}_{comb}~{\widehat{\mathbf{J}}_{comb}^T})}
	\end{align}
	where $det(\cdot)$ denotes determinant of a matrix. 
	
	This quantity measures the difficulty of moving the end-effector in a certain direction for conventional redundant rigid link robots.
	Although we cannot guarantee the estimated combined Jacobian behaves as the true Jacobian, the sudden variations in this measure can provide information regarding the CM response when interacting with environment.
	
    \section{ Results }\label{sec:results}
	
	Fig. \ref{fig:summary} summarizes type of experiments we performed to evaluate the DD-PMI algorithm on the mentioned CMs.
	As shown, to demonstrate versatility and capability of our method, we implemented our algorithm in position control of the CM in free and obstructed environments (with various geometry and stiffness) as well as manipulation of a deformable object with unknown behavior under two imposed boundary condition.In addition, to investigate the viability of the DD-PMI algorithm on \textit{ex-vivo} tissue, we performed an experiment on a lamb heart. We also performed experiments to evaluate the parameters affecting the convergence of the DD-PMI algorithm.
	The following sections describe these experiments in detail.
	
	\subsection{ End-Effector Position Control In Free Environment }
	
	We evaluated the performance of the DD-PMI framework in controlling the end-effector position of each CM in free environment.
	To this end, we first assigned a feasible desired target point $\bm{\Gamma}_d \in \mathbb{R}^2$ (marked with a red point in Fig. \ref{fig:c1}a and Fig. \ref{fig:c2}a ) in the image plane, and then arbitrarily initialized $\widehat{\mathbf{J}}_{comb}(0)\in \mathbb{R}^{2 \times 3}$ with a matrix of all ones.
	For both robots, we performed several preliminary experiments to obtain the optimal values of the following parameters: $\beta=0.7$, $\Lambda_1=\mathrm{diag}(2,2,2)$ for filtering the PSM's joint angles signals provided by the dVRK system, and $\Lambda_2=\mathrm{diag}(2,2)$ for the visual feedbacks of the image feature point provided by the ECM.
	We also set $\epsilon= 1$ pixel as the error threshold. Of note, we repeated this set of experiments for both robots and with different initial conditions and desired target points.
	
	\subsubsection{5mm EndoWrist DeBakey Forceps:} \label{sec:5mmfree}
	
	\begin{figure}[!t]
		\centering
		\includegraphics[width=.5\linewidth]{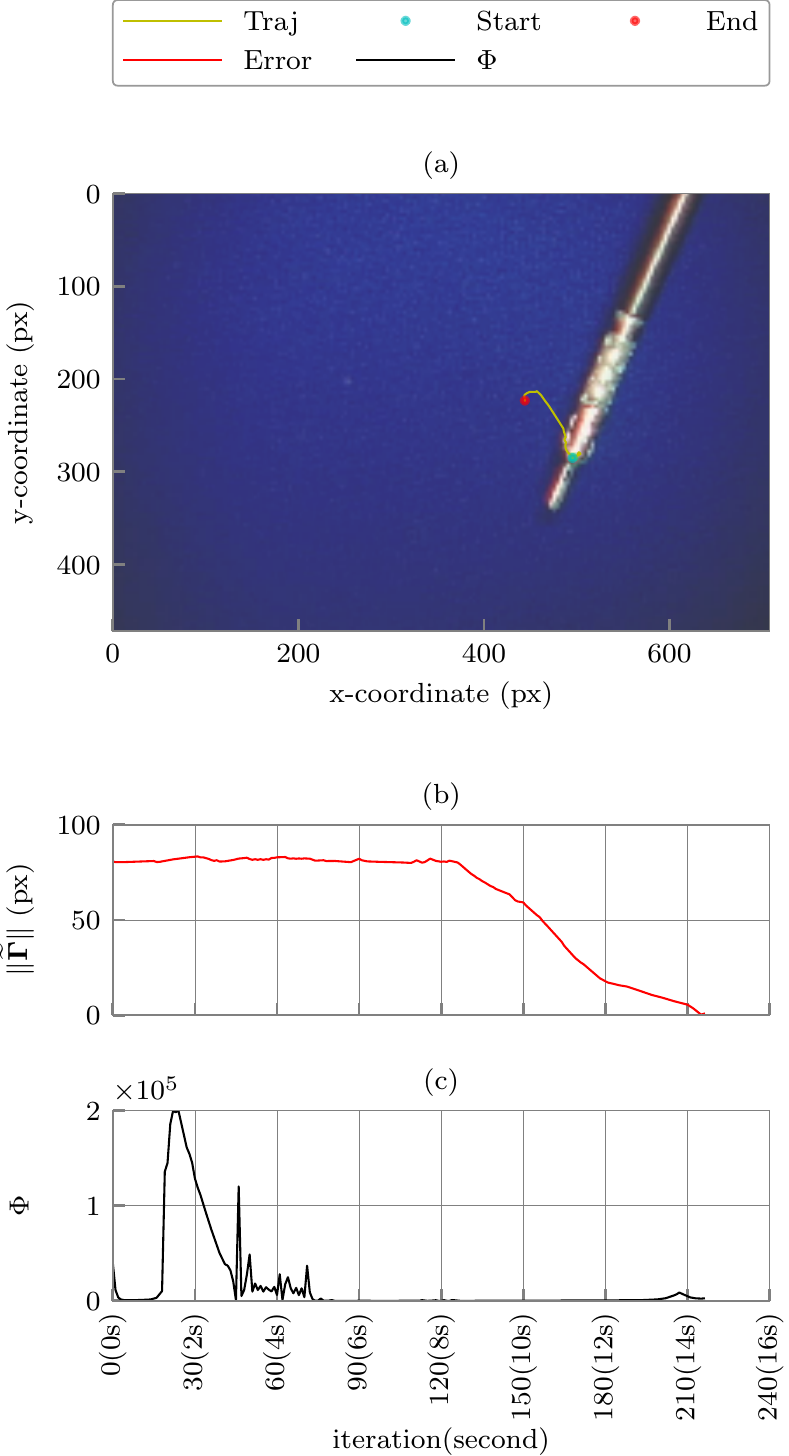}
		\caption{(a) overlay of the traversed trajectory of the 5mm CM's end-effector with respect to the robot's initial configuration during its position control in free environment.
		The green and red points represent the start and end positions of the CM's tip in the image plane, respectively.
		(b) the Euclidean distance error (EDE) between the end-effector position and its desired position.
		(c) the YMM of the estimated Jacobian along the end-effector trajectory.}
		\label{fig:c1}
	\end{figure}
	
	Fig. \ref{fig:c1} demonstrates the traversed trajectory of the end-effector in the image plane (in \textit{pixel}) as well as the EDE between the end-effector position and its desired location in each time instant.
	The Jacobian-related measure (i.e. the YMM of the estimated Jacobian along the end-effector trajectory) is also plotted in this figure.
	As shown in the EDE plot, it takes about 124 steps for the algorithm to identify and learn the Jacobian of the CM considering the initialized Jacobian.
	After this \textit{learning} phase, the robot quickly approaches to the desired target location (\textit{converging} phase).
	Investigation of the YMM measure during the \textit{learning} phase also demonstrate the quick changes in the Jacobian to capture the deformation behavior of the instrument.
	Of note, during the \textit{converging} phase, the rate of changes in the YMM measure drastically decreases, which shows the accuracy of the estimated Jacobian during the \textit{learning} phase. 
	
	\subsubsection{3mm Continuum Manipulator:}
	
	\begin{figure}[!t]
		\centering
		\includegraphics[width=.5\linewidth]{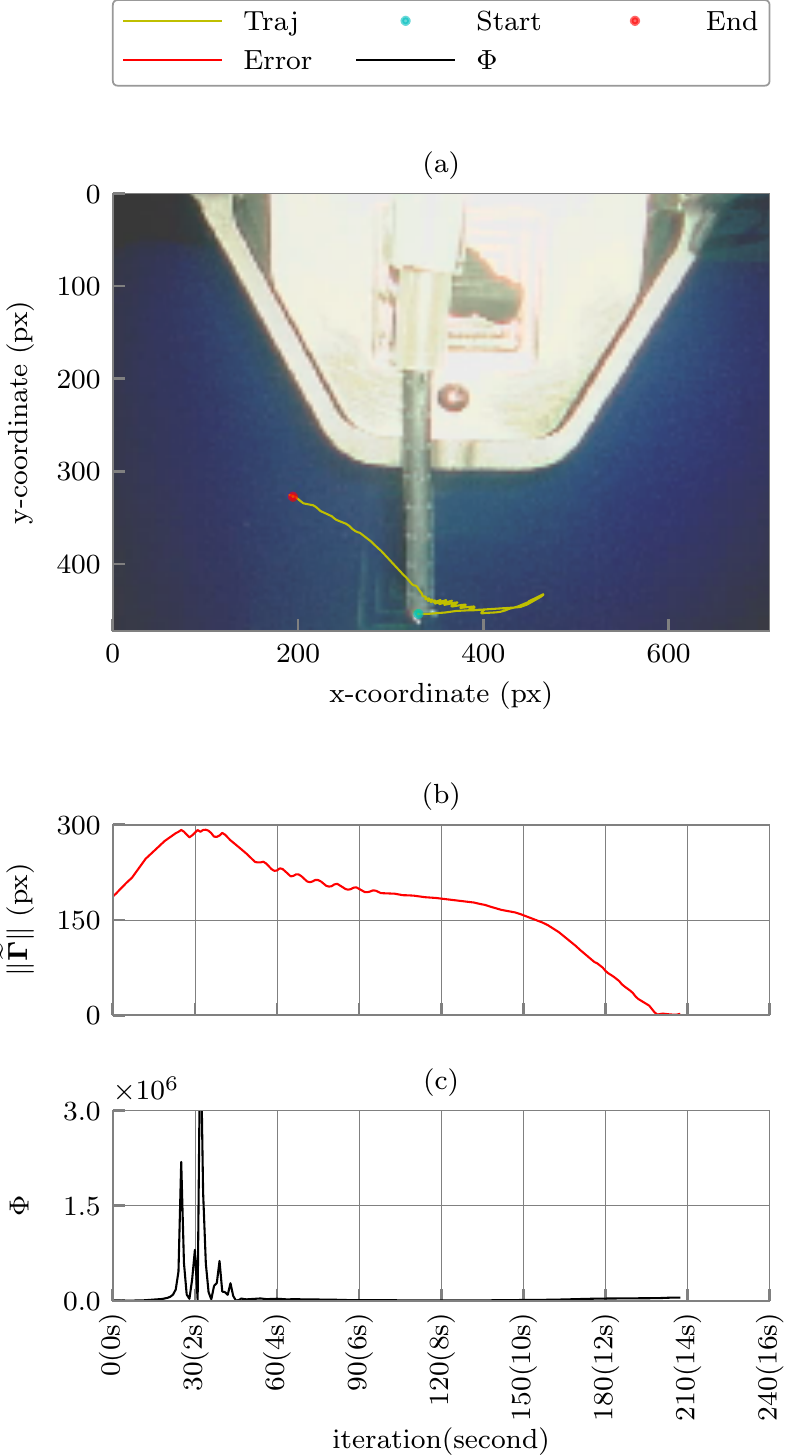}
		\caption{(a) overlay of the traversed trajectory of the 3mm CM's end-effector with respect to the robot's initial configuration during its position control in free environment.
		The green and red points represent the start and end positions of the CM's tip in the image plane, respectively.
		(b) the Euclidean distance error (EDE) between the end-effector position and its desired position.
		(c) the YMM of the estimated Jacobian along the end-effector trajectory.}
		\label{fig:c2}
	\end{figure}
	
	We also repeated this experiment with various initial configurations and desired target points for the 3mm continuum manipulator.
	Fig. \ref{fig:c2}a demonstrates a snapshot of a typical trial as well as the traversed trajectory during the experiment.
	Successful end-effector control of this CM is perceived from the EDE graph besides the trajectory of the end-effector (as shown in Fig. \ref{fig:c2}a-b).
	The YMM measure also represents the behavior of the robot and the update rule during this trajectory (Fig. \ref{fig:c2}c).
	As shown in the EDE plot, in the beginning (steps 0 to 26), due to the inaccurate initialization of the Jacobian, the error increases.
	During this \textit{learning} phase, the CM identifies the true Jacobian of the robot.
	The subsequent steps (steps 27 to 208) correspond to the \textit{converging} phase, when the CM moves toward the target and EDE decreases.
	Of note, during steps 27 to 112, the CM uses the roll motion when moving toward the target.
	The corresponding increase and decrease in the EDE and the YMM measures are due to the imperfections and slack in the roll motion of the CM's actuation mechanism.
	In fact, the Jacobian identifies this imperfection and compensates for it in the mapping between the joint space and end-effector motions.
	
	\subsection{Continuum Manipulator Shape Control}
	
	\begin{figure}[!t]
		\centering
		\includegraphics[width=.5\linewidth]{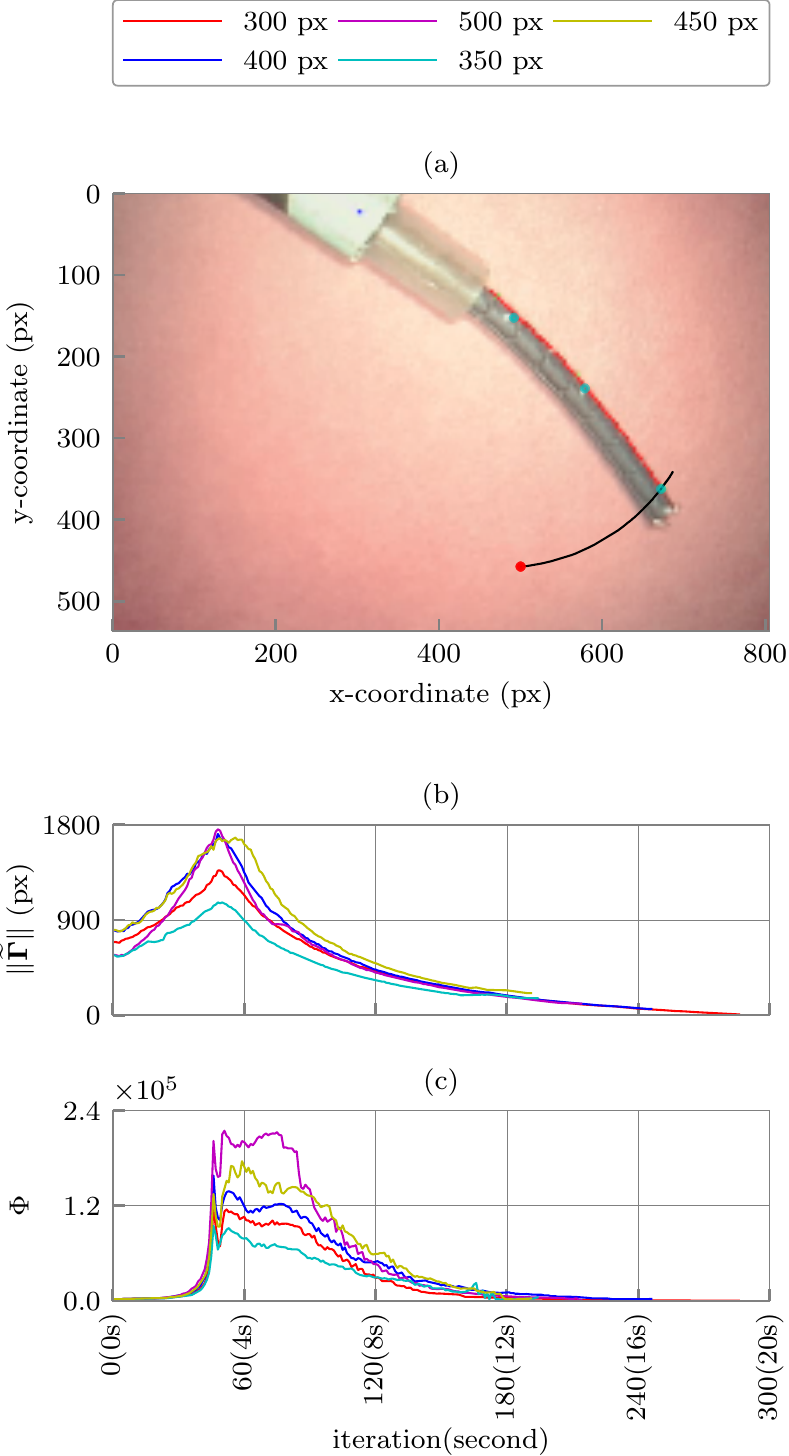}
		\caption{(a) overlay of the traversed trajectory of the 3mm CM's end-effector with respect to the robot's initial configuration during its shape control in free environment with the desired radius of curvature of $\Gamma_d=\kappa^{-1}_d=300$ ${pixel}$.
		The green points represent the location of feature points on the surface of the CM and red point represents the end position of the CM's tip after achieving the desired curvature.
		(b) the Euclidean distance error (EDE) between the end-effector position and its desired position.
		(c) the YMM of the estimated Jacobian along the end-effector trajectory.}
		\label{fig:d1}
	\end{figure}
	
	We used the 3mm CM to evaluate the efficacy of our method in the DD-PMI shape control. We assigned three feature points on the body of the robot (shown by green points in Fig.\ref{fig:d1}a), and used the inverse of the FFV function expressed in (\ref{eq:curvature k}) to fit a constant curvature curve (shown by a red curve in Fig.\ref{fig:d1}a) to these points in each time instant.
	We used 5 desired radius of curvatures- as the desired feature vector $\bm{\Gamma}_d \in \mathbb{R}$- and utilized the same experimental parameters and error threshold (i.e. $\epsilon= 1$ pixel) as the previous set of experiments.
	We arbitrarily initialized $\widehat{\mathbf{J}}_{comb}(0)\in \mathbb{R}^{1 \times 3}$ with a matrix of all ones.
	Fig. \ref{fig:d1}a shows a typical snapshot of the experiment with the desired radius of curvature of $\Gamma_d=\kappa^{-1}_d=300$ ${pixel}$.
	Further, corresponding EDEs and the YMMs are shown in Fig. \ref{fig:d1}b and Fig. \ref{fig:d1}c, respectively.
	According to these figures, for all experiments, it took about 50 steps for the robot to identify and learn the Jacobian and successfully converge to the desired shape.
	It is notable that the length of \textit{learning phase} is dependent on the initialization of $\widehat{\mathbf{J}}_{comb}$.
	Since we started all experiments with the same initial Jacobian, we see almost the same \textit{learning} and \textit{converging} phases. 

	\subsection{End-Effector Position Control In Obstructed Environments } \label{sec:obstructed}
	
	The interaction of the CM with an obstructed environment may affect the CM's kinematics and dynamics as compared to the cases of free motion in the space.
	Further, deformation behavior of the CM is dependent on the location, magnitude, and type (e.g. point load or distributed load) of the contact force as well as the geometry (e.g. flat, concave or convex) and stiffness (e.g. rigid, deformable, or variable) of the unknown obstructed environment.
	Given these uncertainties, therefore, the implementation of a model-based control method in unknown obstructed environments may not be feasible.
	Examples of these situations include interaction of various types of continuum manipulators interacting either with deformable tissues (e.g. in natural orificeTrans-luminal endoscopic surgery \cite{simaan2009design}, and endoscopic submucosal dissection \cite{patel2015evaluation}) or rigid tissues (e.g. in MIS treatment of femur osteonecrosis \cite{alambeigi2017curved} and pelvic osteolysis in hip revision surgery \cite{alambeigi2016design}) with various unknown geometry and stiffness.
	
	\begin{figure*}[!t]
		\centering
		\includegraphics[width=\columnwidth]{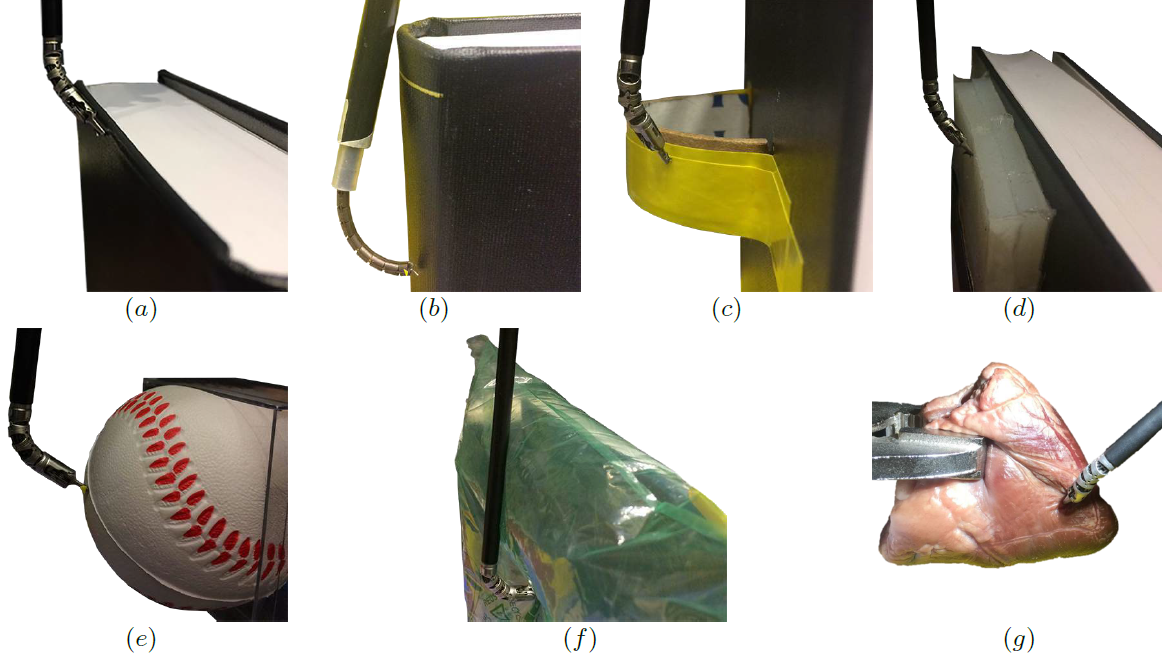}
		\caption{Snapshots of the obstructed environments and type of the CMs used to evaluate the performance of the DD-PMI control framework.
		(a) a rigid and flat environment with 5mm CM.
		(b) a rigid and flat environment with 3mm CM, (c) a rigid and convex environment with 5mm CM.
		(d) a deformable and flat environment with 5mm CM.
		(e) a deformable and convex environment with 5mm CM.
		(f) a variable stiffness and geometry environment with 5mm CM.
		(g) an ex-vivo phantom (lamb heart) with 5mm CM.}
		\label{fig:setups}
	\end{figure*}
	
	Motivated by the mentioned applications, the goal of the experiments was to evaluate the effectiveness of our method for  autonomous maneuver of the CM in the presence of unknown obstacles without prior knowledge of both their material properties and geometries.
	It is worth noting that in the case of PMI control in an obstructed environment, as described in Section \ref{sec:modeling}, deformation behavior of both the CM and the environment have been implicitly included in our formulation.
	Hence, the algorithm learns both deformation Jacobian of the CM and the environment during the interaction phases.
	Accurate learning and estimation of the deformation Jacobian matrix, in fact, affects and assigns the next movement of the CM.
	Convergence of the CM to its predefined error threshold is highly dependent on the learning phase and the estimated deformation Jacobian matrix.
	In order to evaluate the versatility and performance of the DD-PMI framework in learning unknown environments, we considered two rigid and three deformable obstacles with flat, convex, and variable-shape geometries (as shown in Fig. \ref{fig:setups}).
	We chose the target point (marked as red) ensuring the collision of the CM with the obstacle on its trajectory toward the target point.

	In this set of experiments, we used both CMs and arbitrarily initialized $\widehat{\mathbf{J}}_{comb}(0)\in \mathbb{R}^{2 \times 3}$ with a matrix of all ones.
	For both robots, we used the following parameters: $\beta=0.7$, $\Lambda_1=\mathrm{diag}(2,2,2)$ for filtering the PSM's joint angles signals provided by the dVRK system, and $\Lambda_2=\mathrm{diag}(2,2)$ for the visual feedbacks of the image feature point provided by the ECM.
	We also set $\epsilon= 1$ pixel as the error threshold.
	
	\subsubsection{5mm instrument interacting with the rigid and flat environment:}
	
	\begin{figure}[!t]
		\centering
		\includegraphics[width=.5\linewidth]{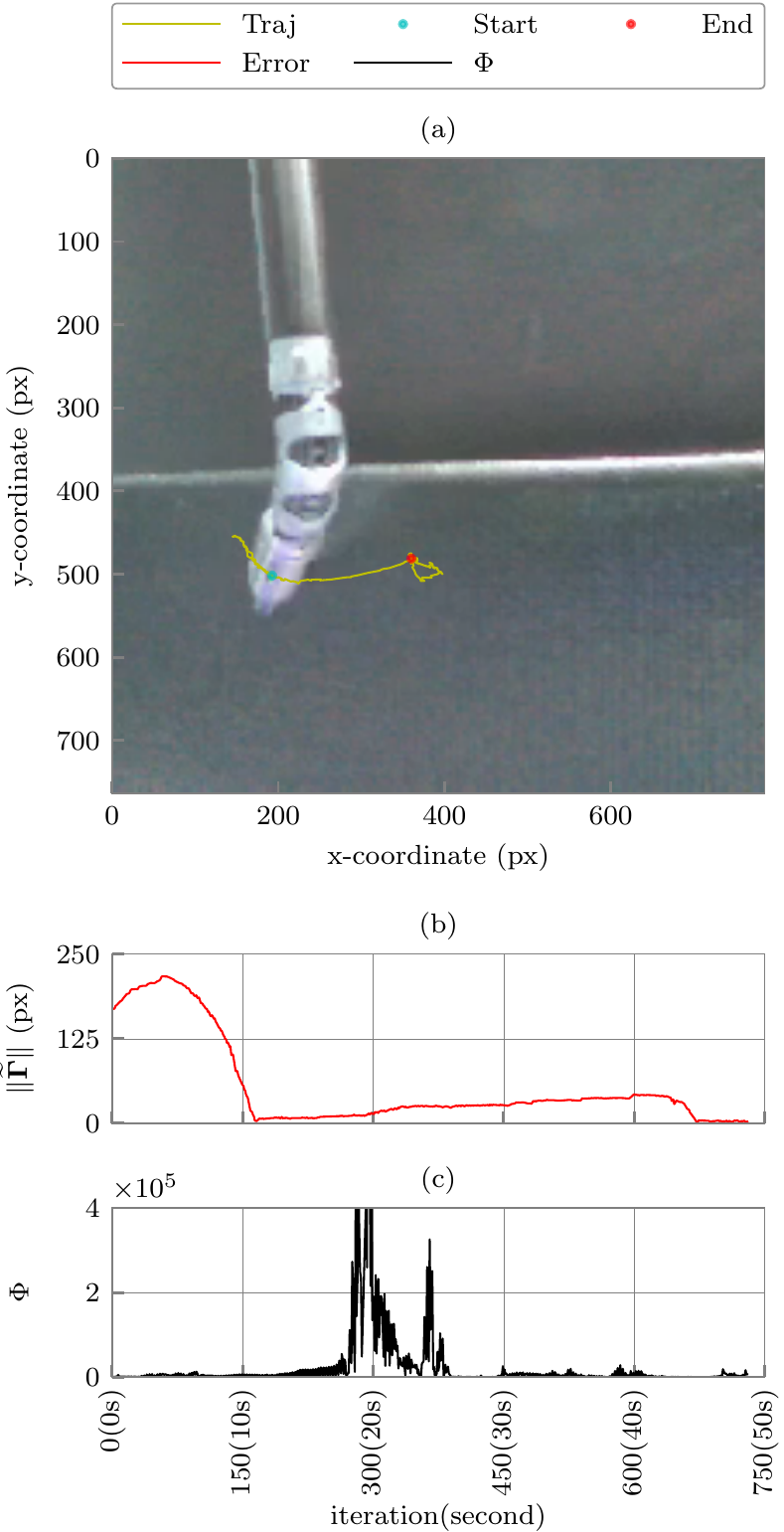}
		\caption{(a) overlay of the traversed trajectory of the 5mm CM's end-effector with respect to the robot's initial configuration during its position control in the flat and rigid obstructed environment.
		The green and red points represent the start and end positions of the CM's tip in the image plane, respectively.
		(b) the Euclidean distance error (EDE) between the end-effector position and its desired position.
		(c) the YMM of the estimated Jacobian along the end-effector trajectory.}
		\label{fig:e1}
	\end{figure}
	
	Fig. \ref{fig:e1}a demonstrates the DD-PMI control of the 5mm instrument and its interaction with a flat rigid obstacle (Fig. \ref{fig:setups}a).
	As shown in Fig. \ref{fig:e1}b, in the first 60 steps, due to the inaccurate initialization of the Jacobian matrix $\widehat{\mathbf{J}}_{comb}(0)$, the Euclidean distance error increased.
	However, after this \textit{learning} phase (step 61 to 165), the EDE drastically decreased (from 216 to 4 pixel) during the (\textit{convergence} phase).
	We call steps 166 to 600 as the \textit{singularity} phase.
	In this phase, regardless of the control inputs (i.e. pull of the actuation cables), the rate of changes in the displacement of the instrument's tip is small due to the interaction with rigid obstacle.
	The vector of the instrument's tip displacement is almost normal to the obstacle surface (i.e. pushing the flat and rigid surface), which results in a small or negligible displacement of the instrument.
	This incorrect movement arises from the lack of knowledge about the environment geometry and stiffness which results in a dynamic change in the CM Jacobian during interaction with the environment.
	Of note, during this phase, the DD-PMI algorithm identifies and learns the new deformation behavior of the CM together with the environment in real time.
	Investigation of the YMM plot (Fig. \ref{fig:e1}c) supports this claim, where multiple peaks in this measure demonstrate the drastic changes in the Jacobian matrix in order to identify the environment and move the CM out from the singularity situation.
	The CM pushes and releases the obstacle multiple times (as can be seen in the provided media), to identify and learn the deformation behavior of environment and move out of the singularity configurations.
	After step 600, the EDE decreased and the robot moved out from the singularity phase and completed the task (\textit{convergence phase}).
	
	An important information that can be inferred from the YMM figure (Fig. \ref{fig:e1}c) is detection of the interaction phase between the CM and environment.
	Investigation of this figure shows that step 35 is the beginning of the interaction between instrument and the flat obstacle.
	This interaction with obstacle can be identified by comparing the YMM of this experiment with other obstacle free experiments.
	As we can see, due to the texture and friction of the obstacle surface, the frequency of the changes in the YMM is higher than the obstacle-free cases.
	This is due to the fact that friction and interaction forces between the instrument and the environment dynamically changing the deformation behavior of the CM, while in the no interaction cases, these forces do not exist and therefore these rapid changes are not detected.
	In fact, the surface texture has also caused small variations in the traversed trajectory of the CM as well as its EDE (as shown in Fig. \ref{fig:e1}b).
	
	\subsubsection{3mm CM interacting with the rigid and flat environment:}
	
	\begin{figure}[!t]
		\centering
		\includegraphics[width=.5\linewidth]{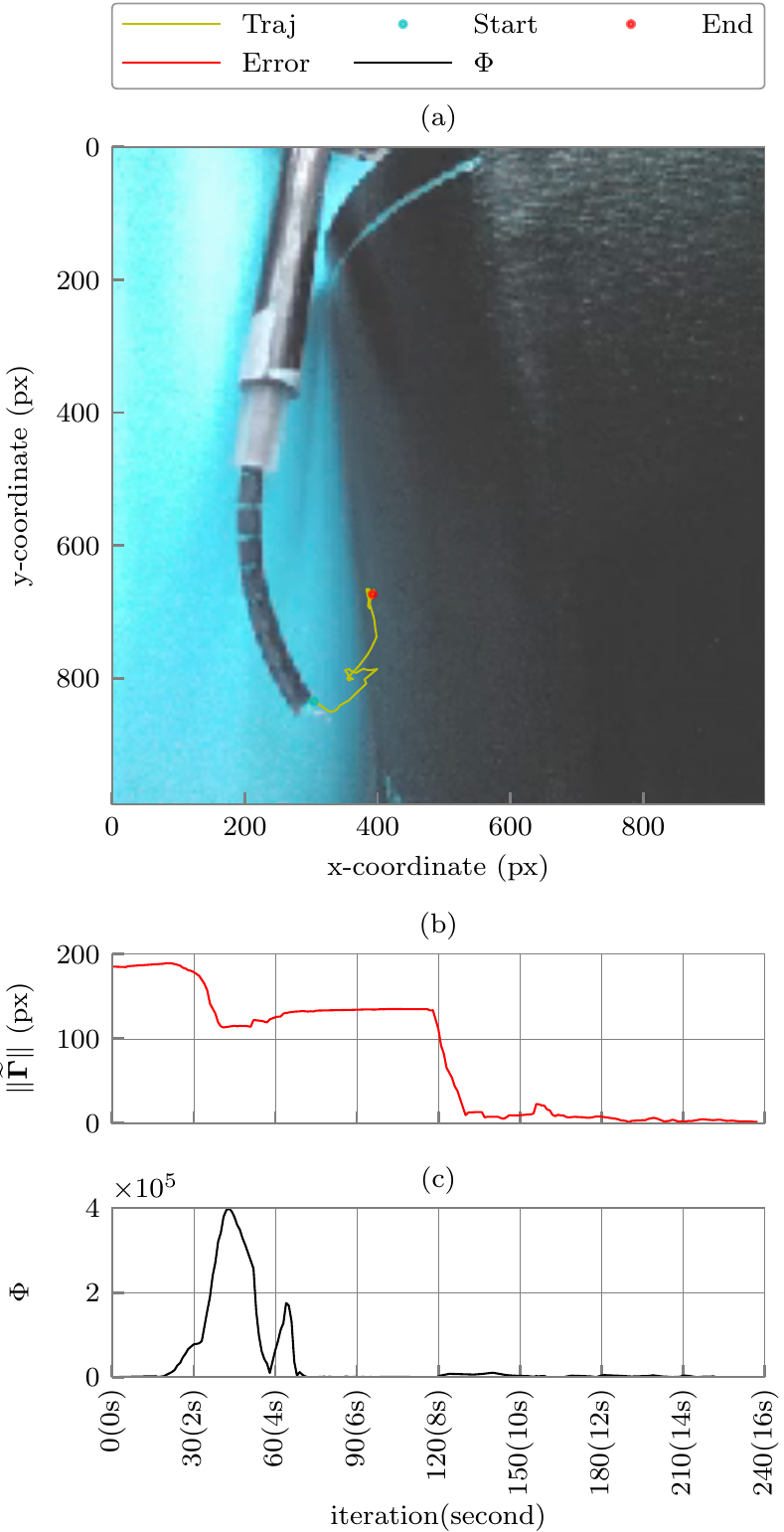}
		\caption{(a) overlay of the traversed trajectory of the 3mm CM's end-effector with respect to the robot's initial configuration during its position control in the flat and rigid obstructed environment.
		The green and red points represent the start and end positions of the CM's tip in the image plane, respectively.
		(b) the Euclidean distance error (EDE) between the end-effector position and its desired position.
		(c) the YMM of the estimated Jacobian along the end-effector trajectory.}
		\label{fig:e2}
	\end{figure}
	
	We repeated the previous experiment with the 3mm CM to study the performance of the DD-PMI method on this robot.
	Fig.\ref{fig:e2} demonstrates the traversed trajectory of this CM with respect to the location of the obstacle in the environment as well as the corresponding EDE and YMM during interaction with this flat and rigid obstacle.
	As shown in Fig. \ref{fig:e2}b, the first two phases (i.e. \textit{learning} and \textit{converging}) can be inferred between steps 0 to 22 and 23 to 42, respectively.
	After these phases, a \textit{singularity} phase occurs (steps 43 to 119).
	Despite the previous experiment which the singularity happened during the interaction of the instrument with the rigid obstacle, here the source of singularity is not solely the collision.
	As mentioned, the 3mm CM has slack in its actuation mechanism, which means despite the movement of its actuation cables, the CM may not bend.
	The disparity between the actuating inputs and displacements can be inferred as another type of singularity.
	The YMM plot during this phase demonstrates the change in the Jacobian to learn the Jacobian and move toward the target point (Fig. \ref{fig:e2}b).
	Steps 120 to 146 corresponds to the \textit{converging} phase, where the EDE decreases quickly before the CM collides with the obstacle.
	Similar to the previous case, a short \textit{singularity} phase occurs due to interaction of the CM with obstacle (steps 147 to 156), where the EDE does not change significantly.
	At step 157, the CM moves away from the singularity and towards the desired target point.
	
	\subsubsection{5mm instrument interacting with the rigid and convex environment:} 
	
	\begin{figure}[!t]
		\centering
		\includegraphics[width=.5\linewidth]{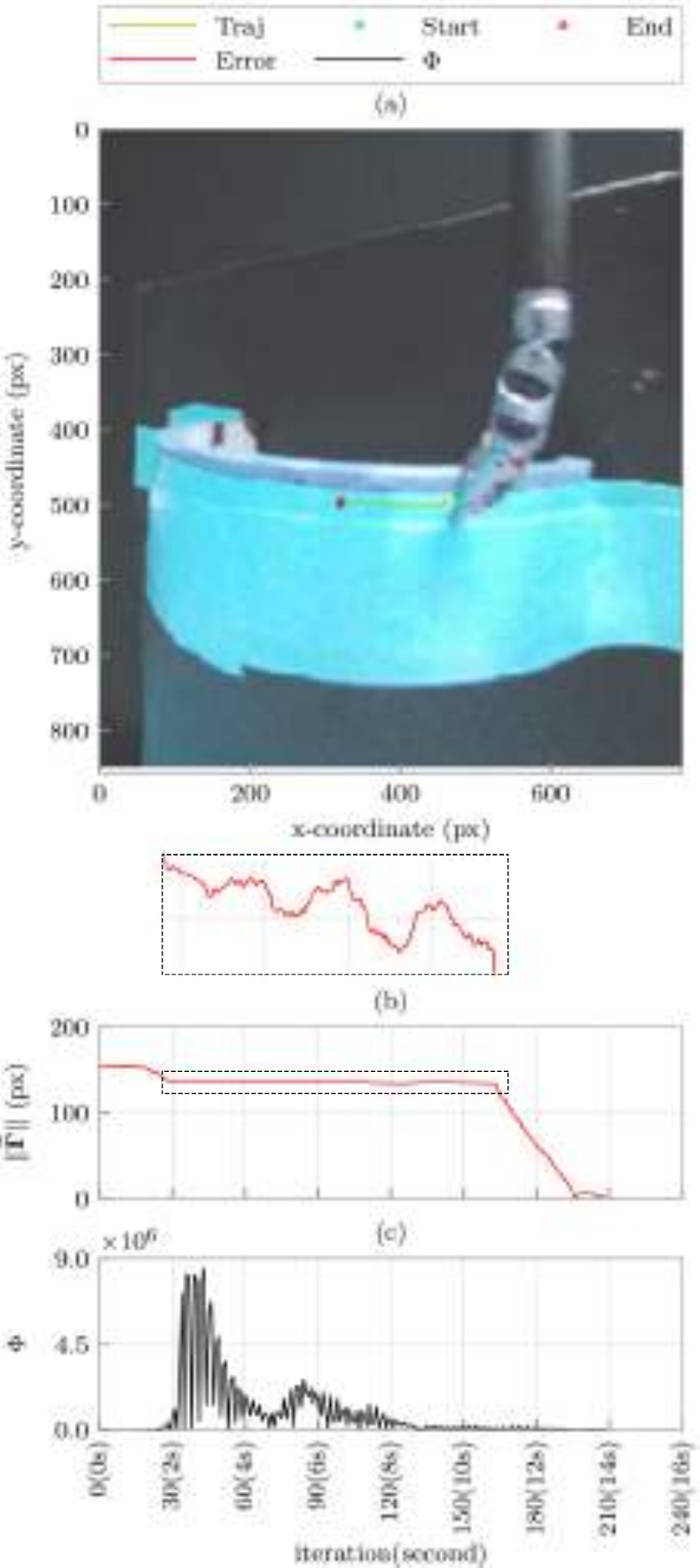}
		\caption{(a) overlay of the traversed trajectory of the 5mm CM's end-effector with respect to the robot's initial configuration during its position control in the convex and rigid obstructed environment.
		The green and red points represent the start and end positions of the CM's tip in the image plane, respectively.
		(b) the Euclidean distance error (EDE) between the end-effector position and its desired position.
		(c) the YMM of the estimated Jacobian along the end-effector trajectory.}
		\label{fig:e3}
	\end{figure}
	
	We selected a convex geometry (Fig. \ref{fig:setups}c) to study our estimation method in the case of interaction of the CM with 1) a non-flat geometry, and 2) an unknown stiff environment with local deformation.

	Fig. \ref{fig:e3}a illustrates the traversed trajectory of the 5mm instrument with respect to the location of the obstacle in the environment as well as the corresponding EDE and YMM during interaction with the convex rigid obstacle.
	Fig. \ref{fig:e3}b demonstrates three introduced phases of \textit{learning}, \textit{singularity}, and \textit{convergence}.
	Steps 0 to 18 corresponds to the \textit{learning} phase, where the EDE increases.
	Subsequently, we see \textit{convergence} towards the target (steps 18 to 30).
	Similar to the previous experiment, due to the interaction with the convex obstacle, the \textit{singularity} phase occurs and lasts for 134 steps.
	During this phase, the robot pushes and releases the obstacle multiple times, as can be seen in the YMM and EDE plots (Fig. \ref{fig:e3}a-b).
	The oscillations in the EDE plot are due to the local deformations of the obstacle during push and release sequences.
	The corresponding changes are also visible through the YMM plot, where quick and oscillating changes in this measure denote the DD-PMI control effort to identify the new deformation behavior during this push-release singularity phase and to move away from this singularity.
	After this phase, the robot moves toward the desired target point in 33 iterations (\textit{converging} phase).
	
	\subsubsection{5mm instrument interacting with the deformable and flat environment:} \label{sec:deformableflat}
	
	\begin{figure}[!t]
		\centering
		\includegraphics[width=.5\linewidth]{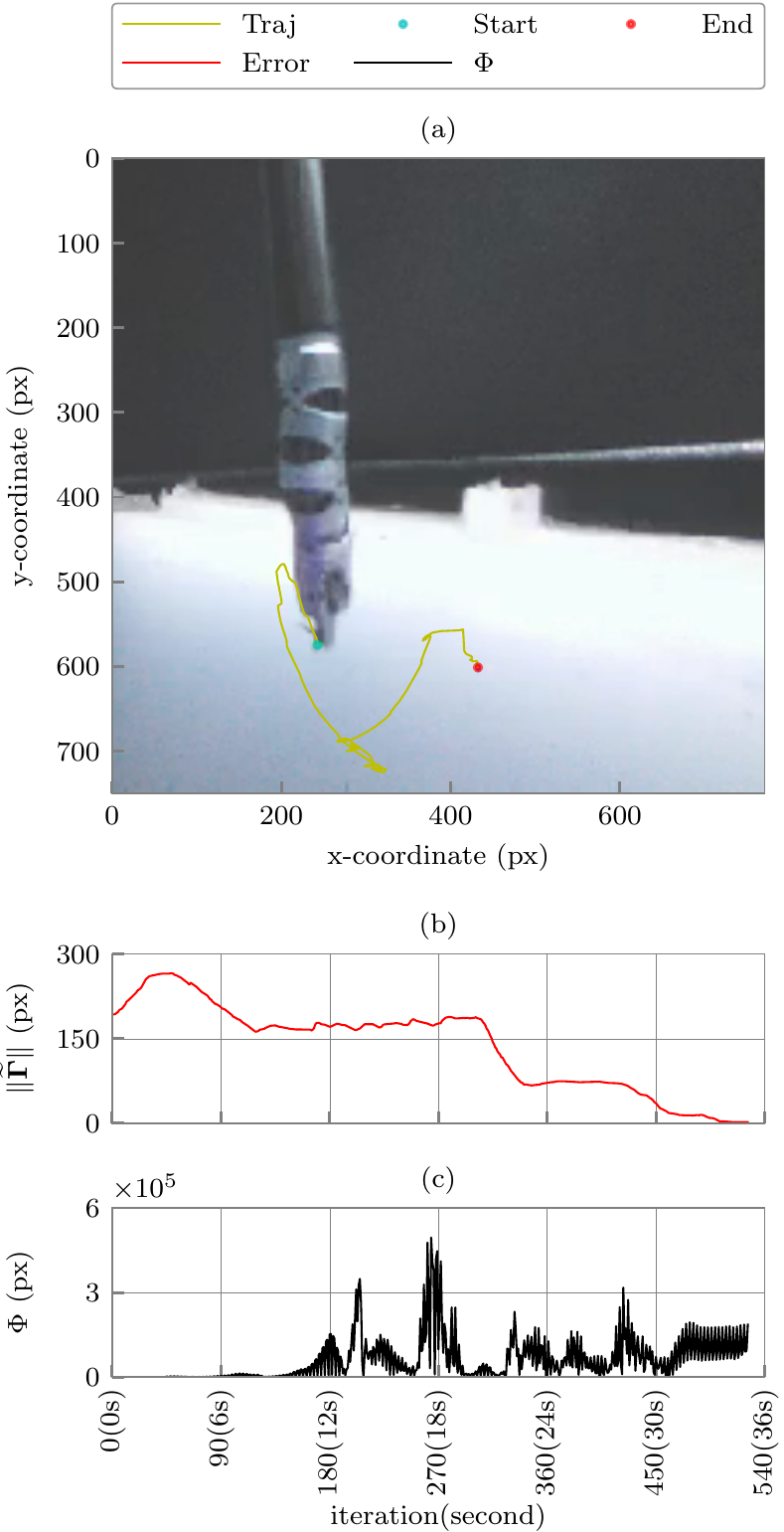}
		\caption{(a) overlay of the traversed trajectory of the 5mm CM's end-effector with respect to the robot's initial configuration during its position control in the flat and deformable obstructed environment.
		The green and red points represent the start and end positions of the CM's tip in the image plane, respectively.
		(b) the Euclidean distance error (EDE) between the end-effector position and its desired position.
		(c) the YMM of the estimated Jacobian along the end-effector trajectory.}
		\label{fig:e4}
	\end{figure}
	
	For this experiment, we used the platinum-catalyzed silicon (Ecoflex 00-50, Smooth-On, Inc.), as a soft, tear resistant, and stretchy material with uniform physical properties, to make a deformable rectangular obstacle ($100 mm\times70 mm\times10 mm$).
	Fig. \ref{fig:e4}a demonstrates the traversed trajectory of the 5 mm instrument in interaction with this obstacle toward approaching the desired target point.
	Further, Fig. \ref{fig:e4}b and \ref{fig:e4}c demonstrate the corresponding EDE and YMM plots of this experiment, respectively.
	The EDE plot shows three \textit{converging} phases (i.e. steps 48 to 118, steps 305 to 346, and steps 427 to 526) separated by two \textit{singularity} phases (i.e. steps 118 to 305 and steps 346 to 427).
	From step 346 to the end of experiment, the instrument interacts with the deformable obstacle.
	During this period, as can be seen in the EDE plot, the CM pushes and releases the obstacle multiple times, which results in a slight decrease and increase in the EDE plot through all this period.
	Further, the corresponding quick changes in the estimated Jacobian can be observed in the YMM plot, which is due to the direct contact between the instrument and the obstacle.
	
	\subsubsection{5mm instrument interacting with the deformable and convex environment:}
	
	\begin{figure}[!t]
		\centering
		\includegraphics[width=.5\linewidth]{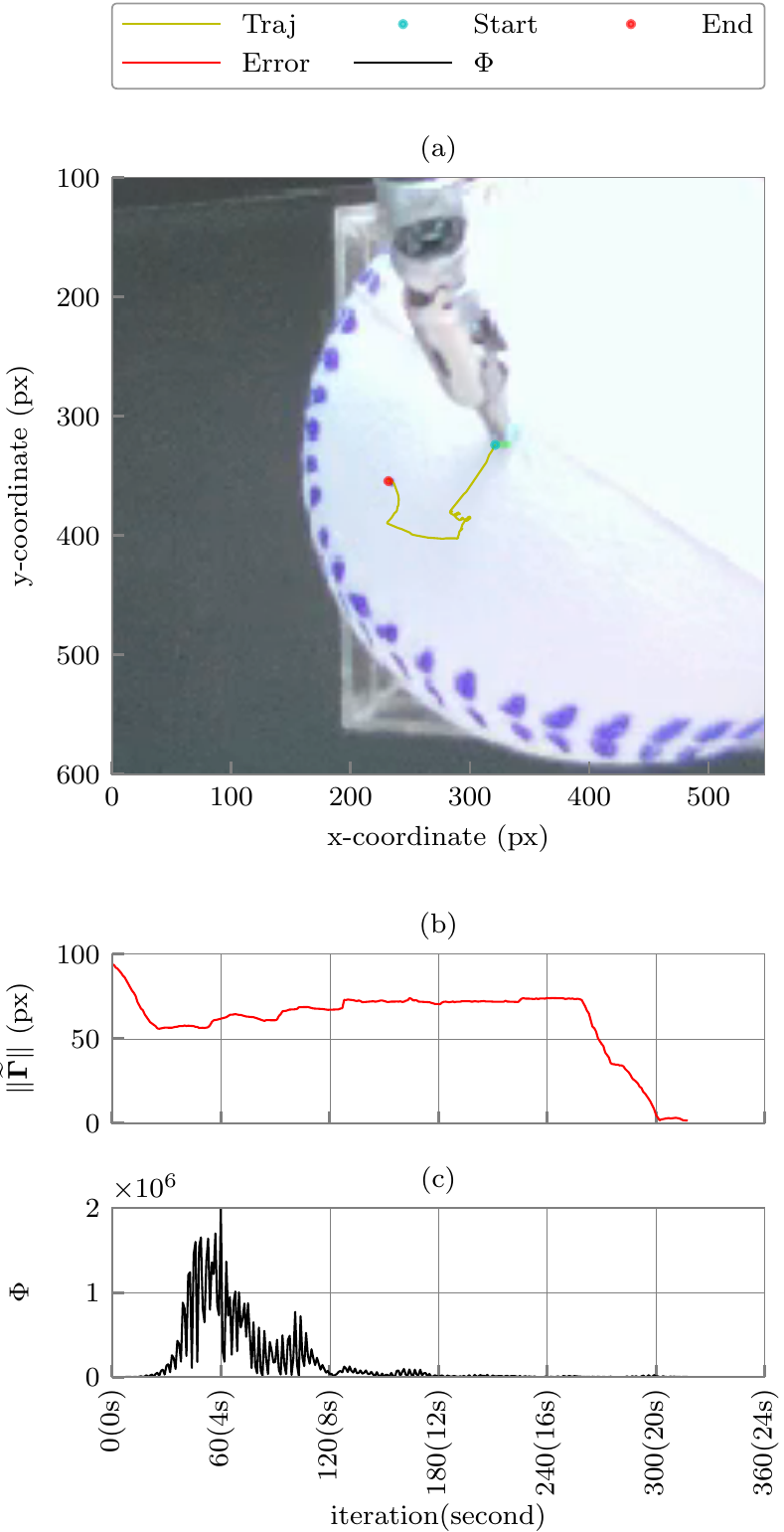}
		\caption{(a) overlay of the traversed trajectory of the 5mm CM's end-effector with respect to the robot's initial configuration during its position control in the convex and deformable obstructed environment.
		The green and red points represent the start and end positions of the CM's tip in the image plane, respectively.
		(b) the Euclidean distance error (EDE) between the end-effector position and its desired position.
		(c) the YMM of the estimated Jacobian along the end-effector trajectory.}
		\label{fig:e5}
	\end{figure}
	
	Fig. \ref{fig:e5}a, demonstrates the interaction of the instrument with a baseball representing a deformable and convex environment.
	In contrast to other cases, we started the experiment when the instrument was pushing the baseball.
	As shown in the EDE plot (Fig. \ref{fig:e5}b), between steps 27 to 260, the instrument pushes and releases the ball multiple times to identify the deformation behavior of the CM while interacting with the environment.
	After this phase, the instrument moves toward the goal by pushing the ball and approaching the desired target.
	Similar to the previous cases, the YMM measure (i.e. Fig. \ref{fig:e5}c) shows a high frequency oscillatory response due to the interaction of the CM with the ball and the contact forces.
	
	\subsubsection{5mm instrument interacting with a variable stiffness and geometry deformable environment:}
	
	\begin{figure}[!t]
		\centering
		\includegraphics[width=.5\linewidth]{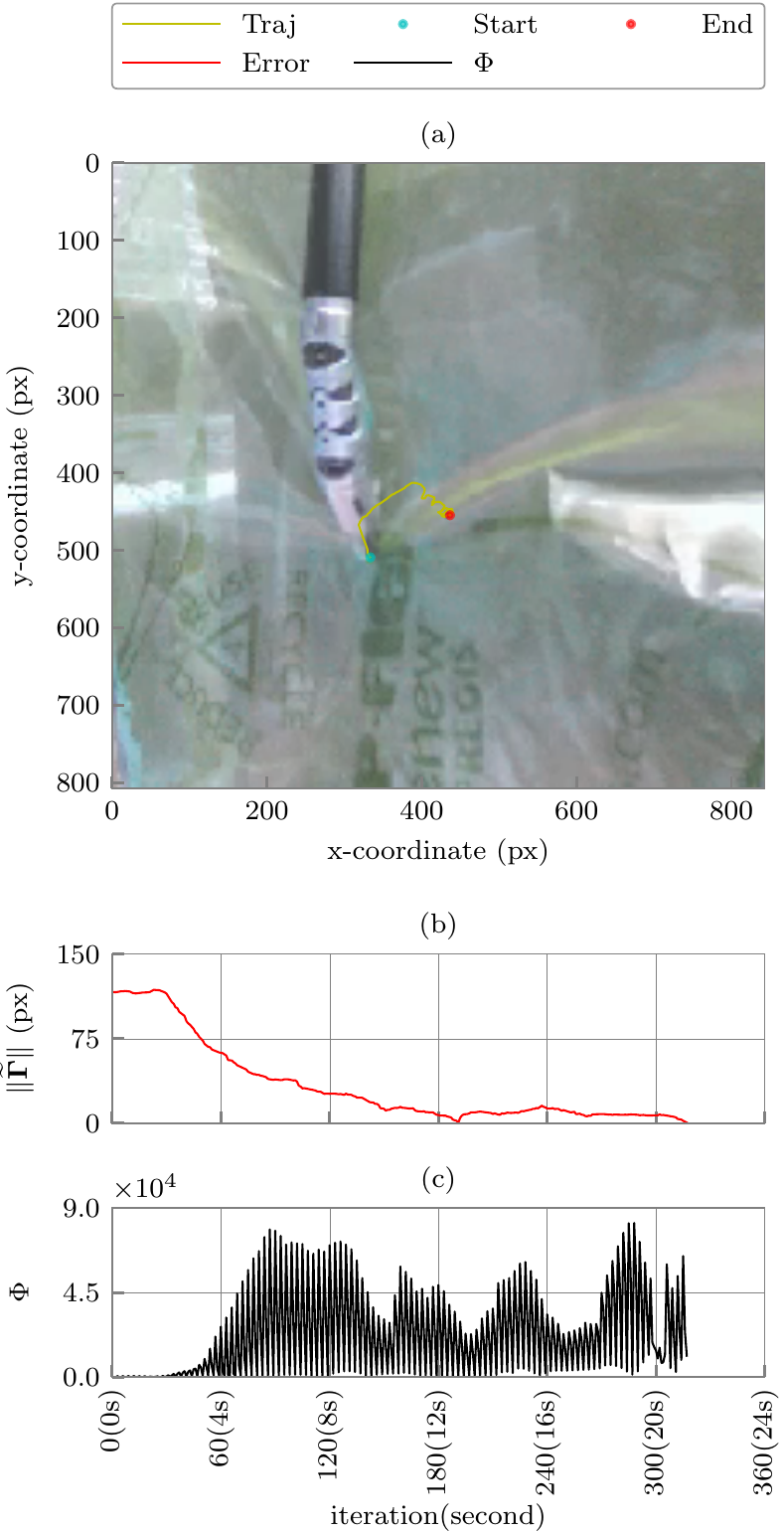}
		\caption{(a) overlay of the traversed trajectory of the 5mm CM's end-effector with respect to the robot's initial configuration during its position control in the variable-stiffness and -geometry obstructed environment.
		The green and red points represent the start and end positions of the CM's tip in the image plane, respectively.
		(b) the Euclidean distance error (EDE) between the end-effector position and its desired position.
		(c) the YMM of the estimated Jacobian along the end-effector trajectory.}
		\label{fig:e6}
	\end{figure}
	
	To investigate the performance of the DD-PMI algorithm in real-time estimation of the deformation behavior of a variable stiffness and geometry environment, we chose an air cushion bag used for packaging (Fig. \ref{fig:e6}a).
	The air cushion bag is very soft such that can be pushed up to a certain depth.
	However, after reaching the mentioned depth, it becomes stiff only in the direction of push- due to the trapped air in the cushion.
	In other words, because of this variable stiffness, a very small movement in the joint space of the robot may result in a big unexpected movement in the Cartesian space.
	Of note, during the experiment, due to the variable trapped air region inside the cushion, the stiff regions vary constantly and the estimator needs to adjust itself in order to predict the deformation behavior by the Jacobian matrix.
	
	Fig. \ref{fig:e6} demonstrates the traversed trajectory of the CM, the EDE, and YMM for this experiment.
	As shown in the YMM plot, despite the other cases, the amplitude and frequency of the YMM oscillatory responses are much higher due to the variable stiffness behavior of the air cushion (Fig. \ref{fig:e6}c).
	According to this plot, at step 55 the air cushion's stiffness begins to drastically change- due to the trapped air- and this variation in the stiffness continues during the experiment.
	Accordingly, we see the corresponding changes in the EDE plot (Fig. \ref{fig:e6}b)- starting at the step 55- which is detectable through sudden increase and decrease in this error.
	The DD-PMI algorithm robustly estimated the deformation behavior in real-time without a prior knowledge and moved toward the target.

    \subsection{Effect of Parameters Initialization on the DD-PMI Algorithm Convergence} \label{sec:convergence}

	As we mentioned in the previous sections, convergence of the DD-PMI algorithm depends on the coefficient $\beta$ defined in (\ref{eq:broyden}) as well as the initialization of the combined Jacobian  $\widehat{\mathbf{J}}_{comb}(0)$.
	To better clarify the effect of these parameters on the convergence of our algorithm, we performed two experiments.
	In these experiments, we used the 5 mm instrument and repeated the experiment described in Section \ref{sec:5mmfree} with similar initial configuration (i.e. end-effector position and orientation) and the desired position.
	
\begin{figure}[!t]
    	\centering
    	\includegraphics[width=.5\linewidth]{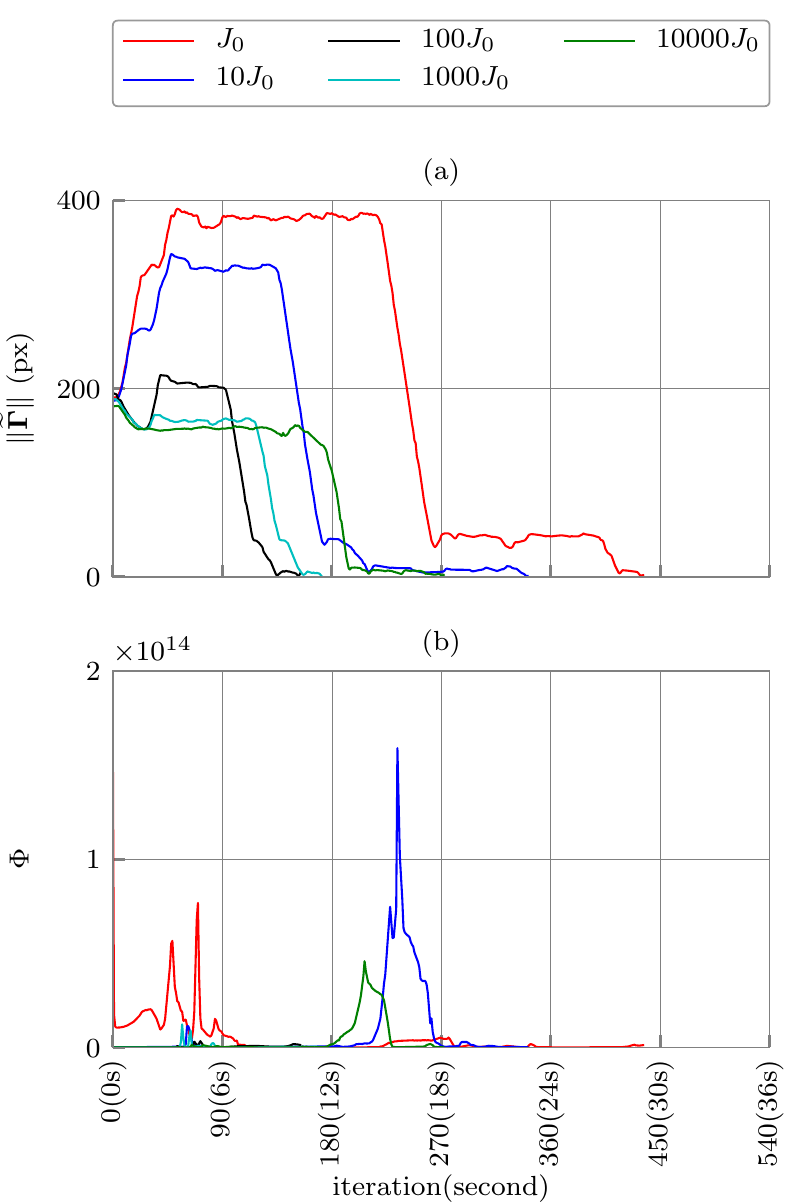}
    	\caption{Comparison of the performance of the DD-PMI algorithm in the case of different initialization of the combined Jacobian matrix.
    	(a) the Euclidean distance error (EDE) between the end-effector position and its desired position;
    	(b) the YMM of the estimated Jacobian along the end-effector trajectory.
    	$J_0$ is a matrix with entries of all ones.}
    	\label{fig_16}
    \end{figure}

    In the first experiment, we fixed  $\beta=0.7$ and arbitrarily varied the initial Jacobian matrix and studied the convergence time and performance (Fig. \ref{fig_16}).
	
    Considering this Figure, we used identical $J_{comb}(0)$ as described in the previous sections and arbitrarily scaled this matrix by a factor of 10, 100, 1000, and 10000.
    As shown in Fig. \ref{fig_16}, two observations can be understood:
    (i) the algorithm convergence does not depend on the matrix initialization;
    (ii) the convergence time can dramatically be affected by this initialization.
    For instance, among the selected initializations, scaling factor of 100 resulted in the fastest convergence time (i.e. $\sim$160 iterations) while the original initialization increased the convergence iterations/times to about three times larger (i.e. $\sim$440 iterations).

\begin{figure}[!t]
    	\centering
    	\includegraphics[width=.5\linewidth]{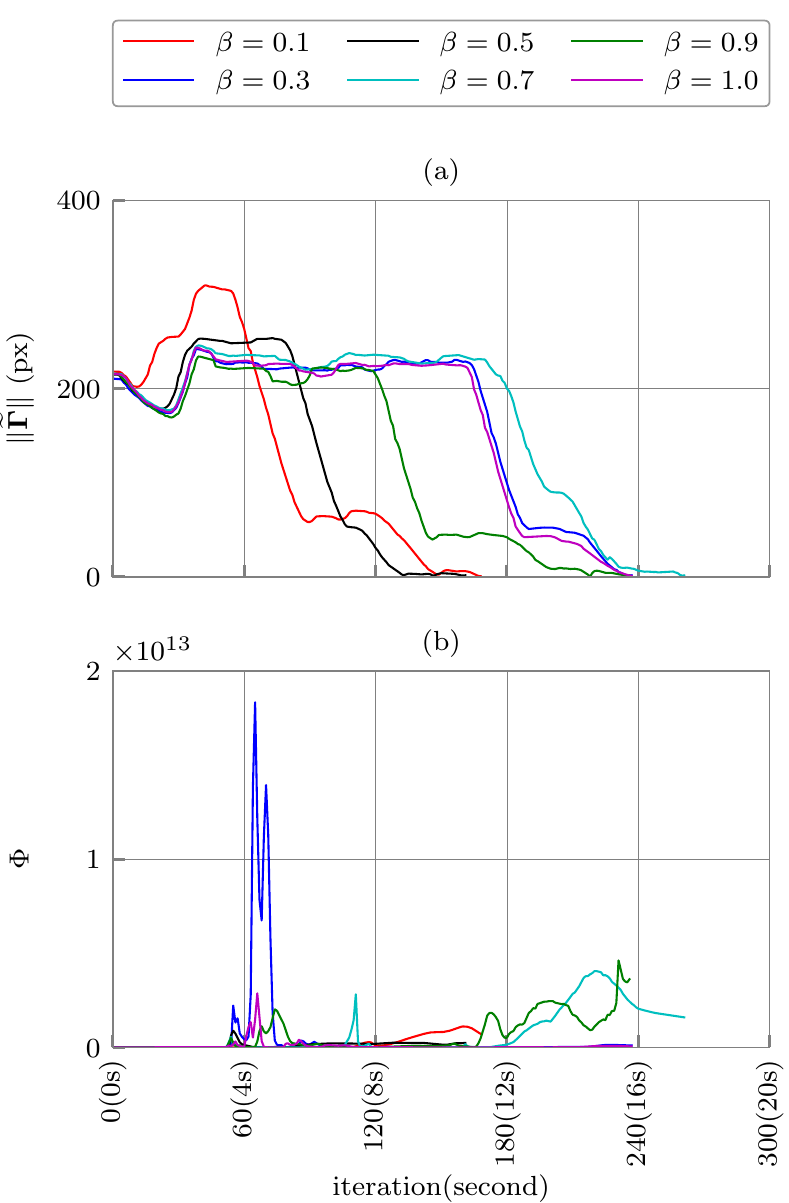}
    	\caption{Comparison of the performance of the DD-PMI algorithm in the case of 5 different $\beta$.
    	(a) the Euclidean distance error (EDE) between the end-effector position and its desired position;
    	(b) the YMM of the estimated Jacobian along the end-effector trajectory.}
    	\label{fig_17}
    \end{figure}

    In the second experiment, we used the best obtained initialization in the previous experiment (i.e. 100 $J_{comb}(0)$) and selected 6 different values for $\beta=\{0.1, 0.3, 0.5, 0.7, 0.9, 1\}$.
    Fig. \ref{fig_17} demonstrates the results of this experiment and the effect of this parameter on the convergence time. As can be observed, similar to the other experiment, for all cases the algorithm converges but with different convergence rates.
    For this particular experiment, $\beta=0.5$ and $\beta=0.7$ had the best and the worst results, respectively.

    \section{Discussion}\label{sec:discussion}
	
	We investigated the performance of the DD-PMI framework for controlling different type of continuum manipulators using three different category of experiments- as summarized in Fig. \ref{fig:summary}.
	We showed that the performance of the DD-PMI was robust to the inaccurate initialization of the estimated deformation Jacobian matrix and initial configuration of the CMs.
	Prior PMI control studies mainly focused on position control of the CMs \cite{yip2014model,melingui2017adaptive,george2017learning}.
	Using the introduced general concept of the feedback feature vector function, however,  we demonstrated that various control objectives can easily be implemented depending upon the available feedback (e.g. S-shape control).
	To show this versatility, we performed both position and C-shape control of the CMs using the DD-PMI framework.
	Further, we successfully showed the scalability of our method  using two different types of CMs designed for medical applications.
	The framework performance is, therefore, independent of the manipulator type and can potentially be extended to other CMs such as concentric tubes \cite{webster2010design} and multi backbone continuum robots \cite{simaan2009design}).
	Furthermore, although we used visual feedback for our control framework, the presented formulation can be adjusted for other types of  sensors (e.g FBG optical or magnetic sensors).
	
	As shown in the results section, generally two phases of \textit{learning} and \textit{convergence} are common in all performed experiments (i.e. free, obstructed, and manipulation experiments).
	In the \textit{learning} phase, the EDE error first increases due to the inaccurate initialization of the deformation Jacobian matrix, then it decreases in the \textit{convergence} phase after learning the Jacobian of the CM or the system of CM/environment.
	In addition to these phases, a \textit{singularity} phase is observed in the experiments pertaining robot and environment interactions.
	In this phase, despite the tension in the actuation cables, no movement in the CM's end-effector is observed due to the interaction of the robot with the obstructed environment (mostly via the robot's tip).
	Here, the change in the EDE measure is very small and usually there is a drastic change in the YMM measure- either frequency or magnitude (e.g. Fig. \ref{fig:e4} and Fig. \ref{fig:e5}).
	In fact, this sudden change in the EDE and YMM measures can also indicate the initial contact instance as well as the contact interval between the CM and the obstacles in the obstructed environment.
	The increase in the EDE measure during the \textit{singularity} phase demonstrates the end of interaction between the CM and environment and moving away from its surface.
	
	The singularity situation may impede the motion of the robot toward the target goal and may cause the robot to get stuck in this situation \cite{yip2016model}.
	To avoid this, we showed that the DD-PMI algorithm could identify and learn in real-time both the CM's and environment's deformation behavior during interaction (using push-release movements) and eventually move out of these situations- by finding an alternate  path to the desired target point.
	Of note, this singularity phase depends on the geometry and stiffness of the unknown obstacles in the obstructed environment.
	For instance, in environments including deformable obstacles (Fig. \ref{fig:e4} and Fig. \ref{fig:e5}) or rigid obstacles with local deformation (Fig. \ref{fig:e3}), the CM may get stuck in various singularity points due to incorrect data-driven estimation of Jacobian- caused by local deformation of the obstacles during interaction with the CM.
	It's worth noting that the estimated deformation Jacobian maps the joint space movements of the robot (e.g. actuation of tendons) to the movement of desired location(s) of the interest on the body of the CM (e.g. CM's tip).
	Hence, local deformation of an environment interacting with the robot is interpreted as a potential path toward the desired target goal by the data-driven algorithm, which in reality is not a possible solution.
	As shown in the previous section, the DD-PMI method could address this singularity situation by accurate estimation of the environment stiffness together with the CM deformation behavior. Of note, this is not guaranteed with the proposed approach by \cite{yip2014model}.
	In Section \ref{sec:modeling}, we discussed and proved that this capability is due to the type of formulation of the deformation Jacobian matrix and features of the Broyden update method (i.e. minimum possible Frobenius norm of the estimated Jacobian with its previous value as well as satisfying the Secant condition).
	To perform similar experiments on flat obstacles, \cite{yip2016model} implemented a hybrid force-position PMI controller to avoid singularity in interaction with the obstacles.
	Here, we showed the  performance of our method in moving out of singularity situations on different stiffnesses and geometries without the need for the contact force information (i.e. direction and magnitude).
	Further, in the case of experiment with the air cushion bag, we demonstrated the performance of the DD-PMI algorithm on a variable-geometry and -stiffness environment without the use of contact force information.
	
	The implemented EDE and YMM measures can also express information about the surface stiffness, where the rising and falling (e.g. Fig. \ref{fig:e4} and Fig. \ref{fig:e5}) in the \textit{singularity} phase are pertained to the local deformation during interaction of the CM with the environment.
	Higher magnitudes of these oscillations can be perceived from a deformable obstructed environment and lower magnitudes may be related to local deformations of a rigid environment.
	In addition, higher frequency in the change of YMM measure demonstrates the rate of variation of the stiffness in the environment (compare Fig. \ref{fig:e6} with other experiments in Section \ref{sec:obstructed}).
	
	The presented results in Section \ref{sec:convergence}, on the effects of Jacobian matrix initialization and coefficient $\beta$, demonstrated the robust performance of the DD-PMI algorithm against random initialization of the mentioned parameters.
	It also showed that optimizing these values can remarkably affect the convergence time, which needs further investigation in our future works.
	Of note, if we include prior knowledge regarding the CM and environment deformation behavior, the convergence rate can be enhanced.
	In all of the experiments performed, we assumed no prior knowledge simulating the worst case scenario.  
	
	\section{Conclusion}\label{sec:conclusion}
	
	In this paper, we presented a versatile data-driven priori-model-independent control framework to manipulate different type of CMs and soft robots in unknown free and obstructed environments without using the interaction force information.
    The main component of our approach is the use of Broyden update rule for estimating the unknown combined Jacobian matrix similar to the techniques used in the visual servoing literature (e.g. \cite{piepmeier2004uncalibrated} ).
    The main idea of using this update rule together with a visual feedback in this study, however, is not merely visual servoing but intelligently utilizing the features of the Broyden update rule to address the limitations of the methods proposed in the literature for model-independent control of CMs.
    Our main contribution is the development of an algorithm that can not only perform visual servoing tasks, but also
    1) works with other type of uncalibrated sensors beyond vision (e.g. FBG \cite{liu2015large,sefati2019fbg} and magnetic tracker sensors \cite{mahvash2011stiffness});
    2) performs shape and end effector position control in both free and obstructed environments without the use of load cells;
    3) avoids singularities due to the end effector interaction with unknown environments (without the use of force feedback) as opposed to \cite{yip2016model}; and
    4) results in the best possible solution to the optimization problem for the model-less control proposed by  \cite{yip2014model}.
    Indeed by using Broyden update rule we have theoretically and experimentally shown all of the above can be simultaneously achieved.
    Furthermore, to the best of our knowledge,  visual servoing literature does not address and deeply investigate the combined problem of uncalibrated visual servoing and model-independent control of a CM.
    We evaluated the DD-PMI framework performance using three different experimental schemes.
	We demonstrated that performance of the DD-PMI method is versatile in terms of independence to the CM type and initial configuration as well as initialization of the estimated deformation Jacobian matrix.
	Further, we demonstrated the efficacy of the DD-PMI approach in manipulation of the CM in 6 various obstructed environments with unknown obstacle geometry and stiffness.
	In addition, the capability of the DD-PMI approach was successfully studied in manipulation of the CM under unknown dynamic disturbances during manipulation of a deformable object. 
    We also demonstrated a successful implementation of the proposed algorithm on an ex-vivo phantom.	

	Some limitations of this study are as follows.
	We limited the shape control of our CMs to only constant curvature C-shape configurations. The success of S-shape control as well as the use of other potential feedback feature vector functions will  need additional investigations.
	The framework was evaluated using visual feedback. Future studies may focus on implementing the DD-PMI algorithm using a CM equipped with FBG for minimally-invasive treatment of pelvic osteolysis \cite{alambeigi2016design,alambeigi2018convex, alambeigi2019use}.
	The study of osteolysis will also require addressing combined PMI control of CMs in integration with robotic manipulators \cite{wilkening2017development}.
	Excessive interaction force between the CM and the tissue during the learning phase of the algorithm may create some safety concerns. One potential remedy to this issue can be the use of FBG sensors for simultaneous shape \cite{sefati2017highly,sefati2016fbg} and force sensing \cite{gonenc2016fbg} for  continuous measurement of interaction forces.
	Another potential extension of this study can be autonomous and semi-autonomous tissue debridement, tissue suturing, and tumor disscetion via indirect manipulation of unknown deformable tissues utilizing CMs \cite{pedram2017autonomous, shin2019autonomous,pedram2019toward, alambeigi2018autonomous, alambeigi2018robust, alambeigi2018toward}.
	The future studies will also include performing in-vivo animal experiments to  mimic more realistic tissue properties and clinical settings.

    As shown in Section \ref{sec:convergence}, convergence of  the  DD-PMI algorithm depends on initialization of the involved parameters. To enhance the algorithm's learning performance, future work must address intelligent selection of the involved parameters before and during the experiments. This may include the use of obtained priori and/or posteriori information of the CM and/or environment. Further, in the case of  detection of an interaction with an unknown environment,  the algorithm may utilize a memory factor to speed up  learning. This feature will especially be useful when  repetitive interactions occur within the same environment.

\bibliographystyle{unsrt}

\end{document}